\documentclass{article}

    \PassOptionsToPackage{numbers, compress}{natbib}
\usepackage{iclr2023_conference, times}
\usepackage{float}
    
\usepackage[utf8]{inputenc} % allow utf-8 input
\usepackage[T1]{fontenc}    % use 8-bit T1 fonts
\usepackage{hyperref}       % hyperlinks
\usepackage{url}            % simple URL typesetting
\usepackage{booktabs}       % professional-quality tables
\usepackage{amsfonts}       % blackboard math symbols
\usepackage{nicefrac}       % compact symbols for 1/2, etc.
\usepackage{microtype}      % microtypography
\usepackage{graphicx}
\usepackage{subfig}
\usepackage{gensymb}
\usepackage{amsthm}
\usepackage{mathrsfs}
\usepackage{amsmath}
\usepackage{amssymb}
\usepackage{mathtools}
\usepackage{wrapfig}
\usepackage{soul}
\usepackage[ruled,vlined]{algorithm2e}
\usepackage{algorithmic}
\usepackage{bbm}

\usepackage[textwidth=1.2in,textsize=tiny]{todonotes}

\newcommand{\ba}[1]{\begin{align}#1\end{align}}

\usepackage{stackengine}

\makeatletter
\newcommand{\distas}[1]{\mathbin{\overset{#1}{\kern\z@\sim}}}%
\newcommand{\BF}[1]{\textbf{#1}}

\newcommand{\beqs}{\vspace{0mm}\begin{eqnarray}}
\newcommand{\eeqs}{\vspace{0mm}\end{eqnarray}}
\newcommand{\barr}{\begin{array}}
\newcommand{\earr}{\end{array}}

\newcommand{\Bmat}[0]{{{\bf B}}}
\newcommand{\Cmat}[0]{{{\bf C}} }

\newcommand{\Tmat}[0]{{{\bf T}}}

\newcommand{\fv}[0]{{\boldsymbol{f}} }

\newcommand{\uv}{\boldsymbol{u}}
\newcommand{\vv}{\boldsymbol{v}}

\newcommand{\xv}{\boldsymbol{x}}

\newcommand{\Sigmamat}[0]{{\boldsymbol{\Sigma}}}

\newcommand{\thetav}{\boldsymbol{\theta}}

\newcommand{\muv}[0]{{\boldsymbol{\mu}}}

\newcommand{\given}{\,|\,}

\usepackage[textwidth=1.2in,textsize=tiny]{todonotes}
\newcommand{\ours}{{PCD}}
\usepackage{pifont}% http://ctan.org/pkg/pifont
\newcommand{\cmark}{\ding{51}}%
\newcommand{\xmark}{\ding{55}}%

\title{%Model-private u
A prototype-oriented clustering for domain shift with source privacy
}
\iclrfinalcopy
\author{Korawat Tanwisuth$^1$, Shujian Zhang$^1$, Pengcheng He$^2$,
\bf{Mingyuan Zhou$^{1}$}\\
$^1$The University of Texas at Austin \quad \quad $^2$Microsoft
\\
\texttt{\{korawat.tanwisuth, szhang19\}@utexas.edu}
\\\texttt{penhe@microsoft.com}
\\\texttt{mingyuan.zhou@mccombs.utexas.edu}
}
\begin{document}

\maketitle

\begin{abstract}
%\xj{no need to specify "unsupervised" for clustering?}
%\xj{nits: no need to mention this paper here, just need to focus on your setting and approach for abstract} \kt{updated}

Unsupervised clustering under domain shift (UCDS) studies how to transfer the knowledge from abundant unlabeled data from multiple source domains to learn the representation of the unlabeled data in a target domain. In this paper, we introduce
Prototype-oriented Clustering with Distillation (PCD) to not only improve the
performance and applicability of existing methods for UCDS, but also address the
concerns on protecting the privacy of both the data and model of the source domains. PCD first constructs a source clustering model by aligning the distributions of prototypes and data. It then distills the knowledge to the target model through cluster labels provided by the source model while simultaneously clustering the target data. Finally, it refines the target model on the target domain data without guidance from the source model. Experiments across multiple benchmarks show the effectiveness and generalizability of our source-private clustering method.

\end{abstract}

 %During target adaptation, the only available information is the unlabeled target data and the source model that we can query to provide cluster labels.
 
 %While unsupervised-domain-adaptation methods can alleviate the domain shift problem, they require labeled data in the source domain.
 %\citet{menapace2020learning} recently created Unsupervised Clustering under Domain Shift (UCDS), a setting where both source and target domains have no labels. \kt{merge the first two sentences} 

% \xj{overall comments: 1. make sure formula is self-contend. 2. not sure about emphasizing proposing a new setting} 
% \kt{TODO:remove all target initialization and rewrite to fix the context}

\section{Introduction}\label{intro}
\vspace{-3mm}

%\sz{please check the Check List at the end}
Supervised learning methods require a tremendous amount of labeled data, limiting their use cases in many situations \citep{adadi2021survey}. By contrast, unsupervised clustering seeks to group similar data points into clusters without labels \citep{hartigan1972direct}. Clustering has become one of the most popular methods in various applications, such as computer vision \citep{coleman1979image, lei2018superpixel, liu2021fusedream, mittal2021comprehensive}, natural language processing \citep{biemann2006chinese,yoon2019compare}, reinforcement learning \citep{mannor2004dynamic, xu2014clustering,ahmadi2021dqre}, and multi-modal learning \citep{hu2019deep, chen2021multimodal}. In many of these applications, data naturally come from multiple sources and may not contain labels since they are expensive to acquire \citep{girshick2014rich, lin2014microsoft}. As an example, medical institutions collaborate to achieve a large and diverse dataset \citep{mojab2020real}. However, this partnership faces privacy and ownership challenges \citep{sheller2020federated}. Across different domains, users may also have varying amounts of resources and data \citep{salehi2019dynamic}. Another example is the inference-as-a-service paradigm, a business scheme where providers serve models trained on multiple sources of data as APIs ($e.g.$, Google AI platforms, Amazon Web Services, GPT-3 \citep{brown2020language}) without giving clients direct access to them. To exploit the rich data from multiple domains for limited-data-and-resource users while also taking into account privacy challenges, one may consider applying methods from Unsupervised Domain Adaptation (UDA)  \citep{shimodaira2000improving, farhadi2008learning, saenko2010adapting}. These methods nonetheless require labeled data in the source domains, making them not applicable in many scenarios.

%Clustering has become one of the most popular toolkit for data analysis and visualization in various applications

% \kt{add ablation on no privacy: use source to initialize target directly}
% \kt{add a paragraph about clustering here}
%\mz{need to explain why in practice people would care about UDA and UCDS, could give some application examples where they can be helpful, especially for UCDS; for example, you can explain how clustering can be useful for data analysis and certain applications and then explain the importance of doing well in  clustering  for a new domain with limited data. also highlight inference as a service and privacy preserving from the beginning}
% \kt{Describe clustering, application for clustering, then clustering with new domain}
%Unsupervised domain adaptation (UDA) \citep{shimodaira2000improving, farhadi2008learning, saenko2010adapting} has gained popularity in recent years since it allows models trained on label-rich source data to adapt to samples from unlabeled target domains, making the learning in the target domain more data-efficient \citep{jiang2022transferability}. This property is especially essential for deep learning models, which typically require a large amount of data for training. Still, one limitation of this transfer learning paradigm is that the source model training requires labeled datasets, which can be expensive to acquire. 

To overcome the assumption of UDA, \citet{menapace2020learning} have recently introduced Unsupervised Clustering under Domain Shift (UCDS), a learning scenario where both the source and target domains have no labels. The goal of this problem setting is to transfer the knowledge from the abundant unlabeled data from multiple source domains to a target domain with limited data. To solve this problem, \citet{menapace2020learning} propose Adaptive Clustering of Images under Domain Shift (ACIDS), a method that uses   an information-theoretic loss \citep{ji2019invariant} for clustering and batch normalization alignment \citep{li2016revisiting} for target adaptation. However, it has two major drawbacks. First, it assumes that we have full access to the source model parameters to initialize the target model before clustering, limiting its use in privacy-sensitive situations where access to the source model is restricted. Second, it requires batch normalization, a specific architectural design of the source model that may not be applicable in some recently proposed state-of-the-art models such as Vision Transformer \citep{dosovitskiy2020image}.

 %, where practitioners first train a clustering model on multiple unlabeled source domain datasets. After obtaining a strong clustering model on the source data, batch normalization alignment \citep{li2016revisiting} and an information-theoretic loss \citep{ji2019invariant} are used to perform adaptation and clustering on the target data. This approach has proven to be effective at learning a target model that clusters data based on semantic content while preserving the privacy of the source data. However, it has two major drawbacks. 

\begin{figure}[t!]
    \vspace{-8mm}
    \centering
        \includegraphics[width=1.0\textwidth]{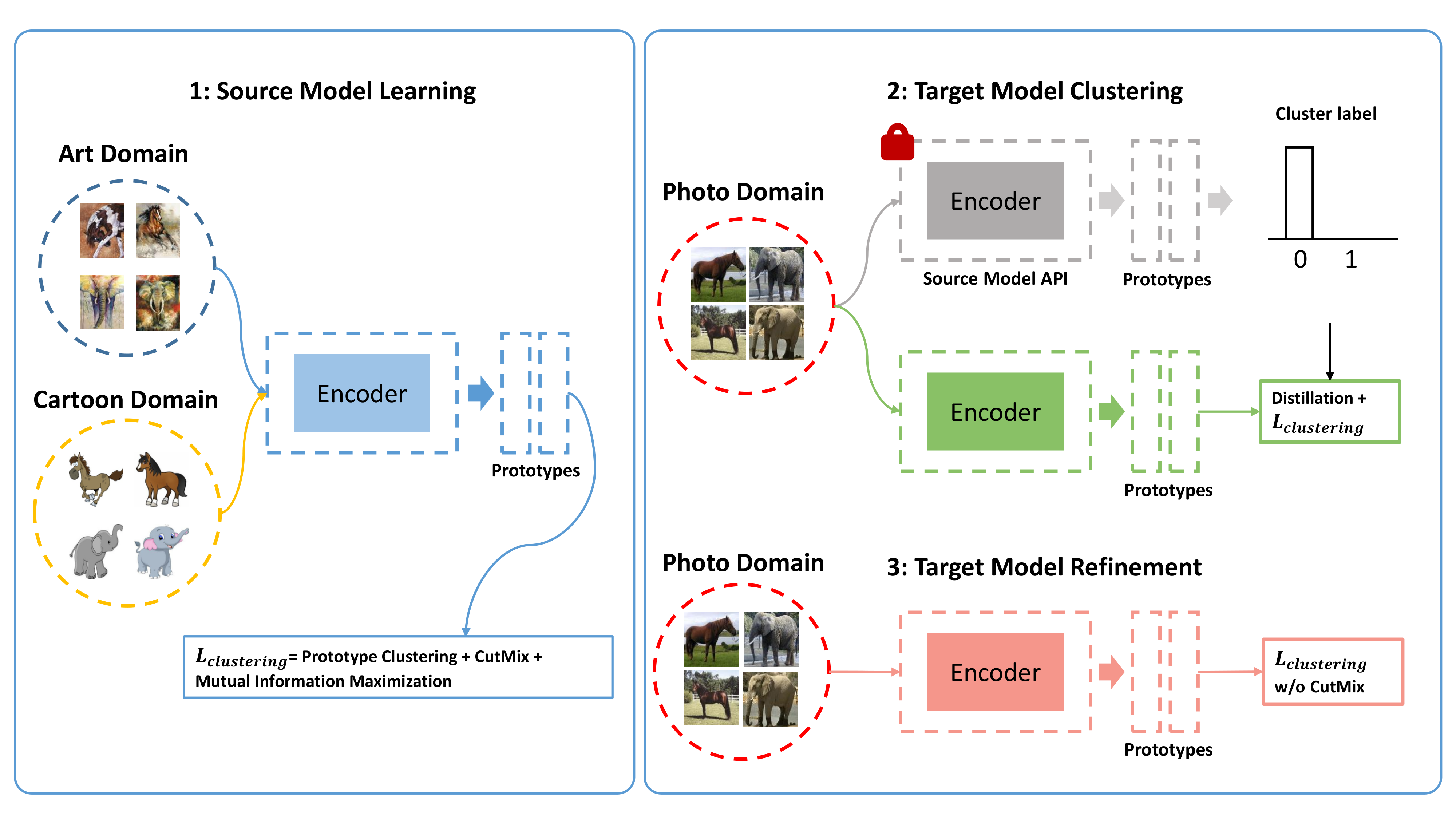}%
        \caption{The illustration of the proposed clustering framework under domain shift and privacy concerns. The semantic content of the source (Art and Cartoon) and target (Photo) data stays the same. However, the bias of the data in each domain leads to a distribution shift. During the adaptation phase, target users are only allowed to query from the source model, protecting the privacy of the source domain information. 
       }
    \vspace{-6mm}

        \label{fig:motivation}%
\end{figure}

In this paper, we consider a more practical problem that is a variant of UCDS (see Table~\ref{tab:transfer_settings}): in addition to the data privacy, we also consider model privacy. Target data owners have no direct access to the source model but can query it to obtain cluster labels during target adaptation. This requirement is important because, given full access to the model, target users or other adversaries may exploit it to recover the source data, jeopardizing source data privacy \cite{chen2019data, luo2020large}. To address this important and challenging problem, we propose Prototype-oriented Clustering with Distillation ({\ours}), a holistic method that consists of three stages. First, we construct a source clustering model from multiple-domain data. To achieve this, we use optimal transport \citep{kantorovich2006translocation, COTFNT} to align the distributions of data and prototypes, as well as a mutual-information maximization to assist the learning of the feature encoder and prototypes \citep{krause2010discriminative, shi2012information, liang2020we}. Second, we use the target cluster assignments provided by the source model to distill the knowledge to the target model while simultaneously clustering the target data. Finally, we perform clustering on the target data alone to further refine the target model. Figure \ref{fig:motivation} illustrates the schematic diagram of our approach.

% \kt{repetition starting from this requirement}
% In this paper, we consider a realistic problem that addresses the challenges faced by UCDS (See Table~\ref{tab:transfer_settings}). We want to transfer the knowledge from a clustering source model trained on multiple source domains to a target domain with limited data. Unlike UCDS, during the adaptation phase, target data owners can only make queries to the source model without directly accessing it. This requirement ensures that we preserve the privacy of not only the source data but also the source model. The privacy of the model in addition to the data is also important. Given full access to the model, target users or other adversaries may exploit it to recover the source data, jeopardizing source data privacy. \kt{CITE}  

% We contrast our problem setting with existing ones in Table \ref{tab:transfer_settings}. 

% \sz{use method, approach instead of framework}

% \sz{the three points in this paragraph and three points in the next paragraph seems overlapping. maybe in the first three three points em} \kt{add citation} \kt{change writing style}

 \ours\ achieves the following benefits. Our approach can be directly applied to the inference-as-a-service paradigm, which is becoming increasingly popular \citep{soifer2019deep}. Many providers currently serve users with API services without sharing direct access to their models. Our method also protects the privacy of both the data and model in the source domains, which is especially critical in practical applications such as healthcare. Moreover, we no longer require the source and target models to share the same architecture, allowing for more flexibility in the training process. Unlike source data owners, target users may have limited resources and cannot afford to train large models. 

Our main contributions include: \textbf{1)} We propose a generalized approach for tackling the problem of data-and-model private unsupervised clustering under domain shift. \ours\ integrates a prototype-oriented clustering algorithm and knowledge distillation into a unified method. Our clustering algorithm synergistically combines optimal transport with the mutual-information objective for prototype and data alignment. \textbf{2)} We verify the effectiveness and general applicability of the proposed method in practical settings: model transfer as well as limited-data and cluster-imbalanced scenarios. \textbf{3)} We provide comprehensive study and experiments on multiple datasets and demonstrate consistent gains over the baselines.

\begin{table}[t!] %htp!]
\caption{Overview of different domain transfer settings.}
 \resizebox{\textwidth}{!}{\begin{tabular}{lcccc}
\toprule  &  Source labels & Target labels  & Source data access &  Source model's parameters access  \\
\midrule
Unsupervised Domain adaptation  & \cmark & \xmark & \cmark & \cmark \\
Source-free Unsupervised Domain Adaptation & \cmark & \xmark & \xmark & \cmark \\
Unsupervised Clustering under Domain Shift & \xmark & \xmark & \xmark & \cmark \\
Ours  & \xmark & \xmark & \xmark & \xmark \\
\bottomrule
\end{tabular}}
\label{tab:transfer_settings}
 \vspace{-4mm}
\end{table}

%and without direct access to the source data and model. During target model adaptation, we can obtain cluster assignments by making queries to the source model. These cluster labels, in turn, allow us to transfer the knowledge from the source domain to the target model. 

%We first introduce the notation and the model before explaining our method in the next section. 

\vspace{-2mm}
\section{Method}
\vspace{-2mm}

To address the clustering problem under domain shift and privacy concerns, we provide a general recipe that consists of three main parts: \textbf{1)} source model learning: learn a transferable model that can guide the target model; \textbf{2)} target model clustering: train a target model with the knowledge from the source model as well as the target data; and \textbf{3)} target model refinement: refine the target model on the target data alone. The resulting strategy, referred to as {\ours}, can effectively solve the clustering problem under domain shift while fully preserving the privacy of the source data and model. We include the pseudocode in Algorithm \ref{alg:pseudo_code} in Appendix \ref{appendix:algorithm}.  
\vspace{-2mm}

\subsection{Background}
\vspace{-2mm}

% In unsupervised clustering under domain shift, we are given $D$ unlabeled datasets from the source domains, denoted as $\{\xv_{dj}^{s} \}_{j=1}^{n_d^s}\sim \mathcal{X}^s^{d}$

In unsupervised clustering under domain shift, we are given $D$ unlabeled datasets from the source domains, denoted as $\mathcal{X}^s = \{ \mathcal{X}^s_d \}_{d=1}^D$ where $\mathcal{X}^{s}_{d} = \{\xv_{dj}^{s} \}_{j=1}^{n_d^s}$ represents a dataset from a source domain $d$ with $n_d^s$ samples. We are also given
an unlabeled dataset from the target domain, denoted as $\mathcal{X}^t=\{\xv_j^{t} \}_{j=1}^{n_t} $ with $n_t$ target samples. There are $K$ underlying clusters in both the source and target domains with similar semantic content, but there is a shift between the source and target data distributions. The clustering model consists of a feature encoder, $F_{\thetav}:\mathcal{X} \rightarrow \mathbb{R}^{d_f}$, parameterized by $\thetav$, and a linear clustering head $C_{\muv}:\mathbb{R}^{d_f} \rightarrow \mathbb{R}^{K}$, parameterized by $\muv$. To simplify the notation, $G=C_{\muv}(F_{\thetav}(\cdot))$ will denote the composition of the feature encoder and linear clustering head. We denote $G^s$ and $G^t$ as the source and target models, respectively.  The goal is to learn a model that can discover the underlying clusters of target samples under domain shift. Although the existing approach by \citet{menapace2020learning} can achieve this objective, it directly uses $G_s$ to initialize $G_t$, compromising the privacy of the source domain and requiring $G_s$ and $G_t$ to have the same architecture. We now discuss how the different components of our method address these issues.

%We denote  ($\thetav^s$, $\muv^s$) and ($\thetav^t$, $\muv^t$) as the source and target model parameters, respectively.
% While these components are standard in domain adaptation for classification \citep{long2015dan,tzeng2017adversarial, courty2017jdot, liang2020we}, the clustering problem and the restriction on privacy of source data and model call for special considerations in each part. 
\vspace{-2mm}
\subsection{Source model learning}
\vspace{-2mm}

To effectively capture the feature distribution of the source data and avoid clustering based on domain information, we propose a clustering algorithm that consists of three components: prototype-oriented clustering, mutual-information maximization, and regularization via CutMix. The first two components help capture the feature distribution, while the last one curtails clustering based on domain information.

%simple strategy that is compatible with modern deep learning models \sz{is it necessary to emphasize the compatible with modern deep learning?}. Our
 %\sz{maybe rewrite a new sentence for this? for example, To effectively capture the features from the source domain, we propose an effective strategy that is compatible with modern deep learning models.} 
\vspace{-2mm}

\subsubsection{Prototype-oriented clustering}
\vspace{-2mm}

%Add hierarchical clustering here. domain invariant feature but dependent proportion in each dataset.
Our goal is to learn global representations of prototypes that capture the source data distributions and a feature encoder that maps the data from different domains to the prototypes. In our model, we have a linear clustering head, $C_{\muv} = [\muv_1, \muv_2, \ldots, \muv_K] \in   \mathbb{R}^{ d_{f}\times K}$, where $d_f$ denotes the dimension of both the prototype and the output of the feature encoder. The vector $\muv_k$ represents a prototype of the $k$th cluster in the latent space. To discover the underlying clusters, we want to align the distribution of the global prototypes with the distribution of the feature representations in each domain.

We represent the distribution of the feature in each domain using the empirical distribution which is expressed as: $P_d = \sum_{j=1}^{n_d^s}\frac{1}{n_d^s}\delta_{\fv_{dj}^s}$ where $\fv_{dj}^s= F^s_{\thetav}(\xv^{s}_{dj})$ denotes the output of the feature encoder. While we use a set of global prototypes to learn domain-invariant representations, we carefully construct the distribution of prototypes in each domain such that the prototypes can align well with the data. Since the proportion of clusters in each domain may vary, we consider the domain-specific distribution of prototypes, $Q_d$, which is defined as: $Q_d = \sum_{k=1}^K\Bmat_{dk} \delta_{\mu_k}$, where $\Bmat_{dk}$ denotes the proportion of cluster $k$ in domain $d$ ($\Bmat_{dk} \geq 0$ and $\sum_{k=1}^K\Bmat_{dk}=1\  \forall d$). We emphasize here that the prototypes are shared across different domains, but the proportion of the prototypes is domain-specific.

%\xj{I lost the logic here}
 To align the distributions of prototypes and data, we want to quantify their difference. A principled way to compare two discrete distributions is to consider the optimal transport problem \citep{kantorovich2006translocation, COTFNT, zhang2021alignment}. Thus, we consider the entropic regularized optimal transport formulation \citep{cuturi2013sinkhorn} that is defined as:
\ba{OT(P_d, Q_d) = \underset{\Tmat_d \in \Pi(\uv, \vv)}{\min} \mathrm{Tr}((\Tmat_d)^T\Cmat_d) + \epsilon h(\Tmat_d), \label{eq:transport_problem}}
where $\Cmat_d \in \mathbb{R}_{\geq 0}^{{n_d^s}\times K}$ stands for the transport cost matrix in domain $d$, $\mathrm{Tr}$ denotes the trace operation, $h(\Tmat_d)=-\sum_{j,k}(\Tmat_d)_{jk}\log (\Tmat_d)_{jk}$ is the entropy of the transport plan, $\epsilon$ controls the strength of the regularization term, and $\Tmat_d \in \mathbb{R}_{> 0}^{{n_d^s}\times K}$ is a doubly stochastic matrix in domain $d$ such that $\Pi(\uv,\vv) = \{\Tmat_d | \Tmat_d \mathbf{1} = \uv, \mathbf{1}^T\Tmat_d=\vv\}$. The probability vectors $\uv = \frac{\mathbf{1}}{n_d^s}\in \Sigmamat^{n_d^s}$ and $\vv = \Bmat_{d} \in \Sigmamat^K$, where 
$\Sigmamat^{M}$ stands for the probability simplex of $\mathbb{R}^{M}$,
%$\Sigmamat^{n_d^s}$ and $\Sigmamat^K$ stands for the probability simplex of $\mathbb{R}^{n_d^s}$  and $\mathbb{R}^K$,  
denote the respective probabilities for %distribution 
$P_d$ and $Q_d$. We define the point-wise transport cost $(\Cmat_d)_{jk}$ as the cosine dissimilarity: $(\Cmat_d)_{jk} = 1-\frac{\muv_k^T\fv^s_{dj}}{||\muv_k||\ ||\fv^s_{dj}||}$, where $\fv_{dj}^s= F^s_{\thetav}(\xv^{s}_{dj})$ denotes the output of the feature encoder. The intuition here is that if $(\Cmat_d)_{jk}$ is high, it is less likely for sample $j$ to be transported to cluster $k$.

To summarize, for a fixed $\thetav$ and $\muv$, we can solve Eq. \eqref{eq:transport_problem} to obtain $\Tmat_d$, the probabilities of moving prototypes to data points in each domain. After obtaining the transport plans, we update the parameters of the encoder $\thetav$ and prototypes $\muv$ to minimize the total transport cost for the given transport plan using mini-batch stochastic gradient descent. The final transport loss is expressed as: $ \mathcal{L}_{transport}\left(G^s; \mathcal{X}^s\right) = \frac{1}{D}\sum_{d=1}^DOT(P_d, Q_d)$. The connections of our method with other deep clustering algorithms \citep{caron2018deep, asano2019self} are provided in Appendix \ref{appendix:connection_deepcluster}.

%\mz{We also need to discuss connections and differences to "PCT" and "Representing Mixtures of Word Embeddings with Mixtures of Topic Embeddings"}

%To optimize this loss, we alternate between solving the optimal transport problem in Eq. \eqref{eq:transport_problem} and updating the parameters of the model with $\mathcal{L}_{transport}$.

%To achieve this, we minimize the transport loss between the distribution of the global prototypes $C_{\muv}$ and the distribution of the penultimate feature representations in each domain as these vectors capture high-level semantic content in the samples.

\vspace{-2mm}

\subsubsection{Learning domain-specific cluster proportions}
\vspace{-2mm}

In the previous section, we utilize cluster proportions, $\Bmat_d$, as the marginal constraint when solving the optimal transport problems. Assuming that each cluster contains roughly the same number of samples, we can use a uniform distribution for $\Bmat_d$. However, this assumption is not valid in practice. Since each domain may have different distributions
over the clusters, we propose a way to estimate domain-specific cluster proportions, $\Bmat_d$. To infer these quantities, we first initialize them with a uniform prior over clusters $\Bmat_{dk} = \frac{1}{K}$ and iteratively refine them using an EM-like update \citep{saerens2002adjusting, kang2018deep, alexandari2020maximum}:
\ba{\textstyle \tilde{\Bmat}_{dk}^{l+1}= \frac{1}{M_d}\sum_{j=1}^{M_d}\pi_{\thetav}^l(\muv_{k}\given \fv_{dj}^s), \hspace{2mm}\text{where} \hspace{2mm} \pi_{\thetav}^l(\muv_k\given\fv_{dj}^s)=\frac{\exp(\muv_k^T\fv_{dj}^s)\Bmat^{l}_{dk}}{\sum_{k'=1}^K\exp(\muv_{k'}^T\fv_{dj}^s  )\Bmat^{l}_{dk'}}, \label{eq:proportions_update}}
where $M_d$ stands for the number of samples in domain $d$ in a mini-batch, $\Bmat_{dk}^{l+1}$ refers to the proportion of cluster $k$ in domain $d$ at the $l+1$ th iteration, $\pi_{\thetav}^l(\muv_k\given\fv_{dj}^s)$ denotes the predicted cluster probabilities at the $l$ th iteration, and $\fv_{dj}^s$ indicates the $j$ th feature sample in domain $d$.  
To obtain a reliable estimates of the full dataset, we iteratively update the proportions with $\Bmat_{dk}^{l+1} \leftarrow  \beta^l \Bmat_{dk}^{l} + (1-\beta^l)  \tilde{\Bmat}_{dk}^{l+1},$
where $\beta^l$ follows a cosine learning rate schedule.

%\mz{good to discuss there exists the summary network approach to directly project the B weight}

%\mz{explain why $B_l$ is not used as the prior for $B_{l+1}$}
\vspace{-2mm}

\subsubsection{Global alignment with mutual-information maximization} %TODO: change this to sound like a regularizer?
\vspace{-2mm}

The transport loss introduced in the previous section aligns the local distributions of data and prototypes. To assist the learning of the feature encoder and prototypes on a global level, we utilize the widely-adopted mutual-information objective \citep{krause2010discriminative, shi2012information}. This objective ensures that the feature representations are tightly clustered around each prototype. %while preventing a trivial solution in which all data are clustered around one prototype.
%Since we initialize our encoder with a pre-trained backbone, we expect that the model should be able to extract semantic information in the image. 
%repetitive?
If the data are close to the prototypes, we expect the posterior probabilities to be close to one-hot vectors. To make this more likely, we minimize the entropy of the conditional distribution of cluster labels given the data. However, minimizing this loss alone could lead to a degenerate solution since the model can assign all the samples to one cluster \citep{morerio2017minimal,wu2020entropy}. To prevent such a solution, we maximize the marginal entropy of the cluster label distribution to encourage the average predictions to be close to a uniform distribution. The mutual-information objective is thus expressed as:
\ba{\mathcal{L}_{ mi}\left(G^s; \mathcal{X}^s\right) &=-[H\left(\mathcal{Y}^{s}\right)-H\left(\mathcal{Y}^{s} \mid \mathcal{X}^s\right)]  \nonumber \\ &=-[h\left(\mathbb{E}_{\xv^{s} \in \mathcal{X}^s} G^s\left(\xv^{s}\right)\right)-\mathbb{E}_{\xv^{s} \in \mathcal{X}^s} h\left(G^s\left(\xv^{s}\right)\right)], \label{eq:mi_loss}}
where $H\left(\mathcal{Y}^s\right)$ and $H\left(\mathcal{Y}^s \mid \mathcal{X}^s\right)$ denote the marginal entropy and conditional entropy of the cluster labels $\mathcal{Y}^s$, which are latent variables, respectively and $h(p) = -\sum_i p_i\log p_i$.
To avoid clustering based on domain information, we add the CutMix \citep{yun2019cutmix} regularization, which mixes two samples by interpolating images and labels. Since the data have no labels, the predicted cluster probabilities are utilized as the pseudo-labels. The CutMix regularization is defined as: $\mathcal{L}_{cutmix} = \mathbb{E}_{\xv_{i}^{s}, \xv_{j}^{s} \in \mathcal{X}^s} L(G^s(\tilde{x}), \tilde{y})$, where $L(\cdot, \cdot)$ is the cross-entropy loss and ($\tilde{\xv}$, $\tilde{y}$) are the interpolated samples from the pair $(\xv_i^s, G^s_{*}(\xv_i^s))$ and $(\xv_j^s, G^s_{*}(\xv_j^s))$, with $G^s_{*}$ indicating no gradient optimization. We construct the final objective function to update the prototypes and feature encoder.
\ba{\mathcal{L}_{clustering}(G^s; \mathcal{X}^s) = \mathcal{L}_{transport}(G^s; \mathcal{X}^s)  + \mathcal{L}_{mi}(G^s; \mathcal{X}^s)+ \mathcal{L}_{cutmix}(G^s; \mathcal{X}^s). \label{eq:source_joint_loss}}

\vspace{-2mm}

\subsection{Target model learning}
\vspace{-2mm}

Because of the domain shift, we divide our target model learning into two stages---target model clustering and target model refinement---to ensure that the knowledge transferred from the source domain does not interfere with the learning in the target domain \citep{shu2018dirt}. The first phase aims to transfer the knowledge from the source model to the target model while protecting the privacy of the source domain. The second phase focuses on refining the target model so that target samples are tightly clustered around each prototype. 
\vspace{-2mm}

\subsubsection{Target model clustering}
\vspace{-2mm}

In many practical applications, it is crucial to preserve the privacy of both the source model and data \citep{ziller2020privacy, ziller2021medical}. Thus, directly using the source model to initialize the target model is not ideal. Instead, we consider the practical problem where the source model can only provide a cluster label for each target example. The source model is simply an API, and we have access to neither its architecture nor model parameters. With the predicted cluster assignments given by the source model, we want to learn a well-trained clustering model on the target data. 

%With this framework, the the source and target domain models can be different, allowing for more flexibility in a resource-limited setting.

\BF{Source knowledge transfer with knowledge distillation.}
Given unlabeled target samples, $\{\xv_i^{t} \}_{i=1}^{n_t}$, we can obtain cluster assignments, $ G^s(\xv_i^t)$, through the source model. 
%, where $G^s(\xv_i^t)_k$ is one for the predicted cluster and zero for the rest. We consider hard labels since many popular clustering algorithms may not give soft labels.
Our algorithm can work for both hard and soft labels; however, it is more practical to consider hard labels from the source domain since soft labels may not be available for all models \citep{sanyal2022towards}. Thus, we consider hard label assignments from the source domain in our experiments. To transfer the knowledge from the source to target models, we utilize a knowledge distillation loss \citep{hinton2015distilling} to train the target model to mimic the predicted output from the source. The loss can be formulated as follows: 
$\mathcal{L}_{k d}\left(G^t ; \mathcal{X}^t, G^s\right)=\mathbb{E}_{x^{t} \in \mathcal{X}^t} \mathcal{D}_{k l}\left(G^s(\xv^t)  \| G^t(\xv^{t})\right)$,
where $\mathcal{D}_{kl}(G^s(\xv^t)\| G^t(\xv^t)) = \sum_{k=1}^K G^s(\xv^t)_k\log \frac{G^s(\xv^t)_k}{G^t(\xv^t)_k}$ stands for the Kullback–Leibler divergence between two distributions and $G^t$ is initialized with a pre-trained feature encoder. 
 %\mz{write out the full expression of KL divergence to avoid any confusion}
 
Because of the domain shift, the source model may not always cluster target samples based on their semantic content. Thus, we propose to refine the predicted target assignments using two simple strategies: label smoothing \citep{pereyra2017regularizing,zhang2021knowing, zhang2021learning} and self-ensemble \citep{laine2016temporal,fan2021contextual, kim2021self}. \citet{muller2019does} discover that label smoothing can help the penultimate layer representation form tight clusters, allowing the model to discover underlying clusters more easily. To utilize label smoothing, we interpolate the hard assignments with a uniform distribution to obtain soft labels:
$\hat{y}^{LS}_k = (1-\gamma) G^s(\xv^t)_k + \frac{\gamma}{K}$,
where $\gamma$ is the weight of the uniform distribution. 
As the target model improves, we can leverage its predicted cluster probabilities across different iterations to form a temporal ensemble: $(\hat{y}^t)^{l} \leftarrow \tau (\hat{y}^t)^{l-1} +(1-\tau)G^t(\xv^t)^{l}$, where $\tau$ determines how much weight we give to past assignments, $(\hat{y}_t)^{l-1}$ is the assignment at the $l-1$ th  iteration, and $G^t(\xv^t)^{l}$ is the current assignment. We initialize $(\hat{y}^t)^0$ with the smooth assignments from the source model.  The refined cluster assignments from the source model $\hat{y}^t$ then replaces $G^s(x^t)$ in the distillation loss. Thus, for target model clustering, the training includes the following losses:
$\mathcal{L}_{target\_clustering}(G^t; \mathcal{X}^t, G^s) = \mathbb{E}_{x_{t} \in \mathcal{X}^{t}} \mathcal{D}_{k l}\left(\hat{y}^t \right \| G^t\left(\xv^{t}\right)) + \mathcal{L}_{clustering}(G^t; \mathcal{X}^t). \label{eq:target_init_loss}$
\label{eq:target_clustering}

% \ba{\mathcal{L}_{k d}\left(G^t ; \mathcal{X}^t, G^s\right)=\mathbb{E}_{x_{t} \in \mathcal{X}^t} \mathcal{D}_{k l}\left(\hat{y}^t \right \| G^t\left(\xv^{t}\right)). \label{eq:distillation_loss}}

%\kt{rewrite this to fit the new loss change.}

 %To alleviate this issue, we incorporate the clustering objective in Eq. \eqref{eq:source_joint_loss} to constraint the target model to learn useful variations in its domain. 

%$\mathcal{L}_{target\_init}(G^t; \mathcal{X}^t, G^s) = \mathbb{E}_{x_{t} \in \mathcal{X}^t} \mathcal{D}_{k l}\left(\hat{y}^t \right \| G^t\left(\xv^{t}\right)) + \mathcal{L}_{transport}(G^t; \mathcal{X}^t) + \mathcal{L}_{mi}(G^t; \mathcal{X}^t)+  \mathcal{L}_{cutmix}(G^t; \mathcal{X}^t). \label{eq:target_init_loss}$ 
% While the knowledge from the source domain can benefit the target model, the shift in distribution leads to noisy predicted cluster assignments from the source model. This phenomenon can lead to unconfident predicted cluster probabilities (uniform cluster probabilities). To alleviate this problem, we integrate the mutual-information objective in Eq. \eqref{eq:mi_loss} to encourage the cluster proportions to be well-balanced and the predicted cluster assignments to have low entropy.
% To incorporate pairwise information and regularize the target model, we also incorporate the CutMix regularization. 

%The distillation loss in Eq. \eqref{eq:distillation_loss} only considers point-wise information between source and target model assignments. 
%(TODO) add some explanations here as to why adding more losses
\vspace{-2mm}
\subsubsection{Target model refinement}
\vspace{-2mm}

In the previous section, we use both source and target domain knowledge to learn our clustering model. While the source domain knowledge can assist target domain learning, the bias in distribution due to domain shift could lead the target model to learn noisy domain information from the source model. Similar to the observation by \citet{shu2018dirt}, we find that the target model could benefit from further clustering on the target data alone. We utilize the clustering objective in %Section 
% \ref{sec:consistency_regu}
Eq.~\eqref{eq:source_joint_loss} 
with target data and model as arguments and without the CutMix loss. The CutMix regularization term is not included since there is no source knowledge transfer and the target data come from a single domain. Also, the regularizer makes the predicted probabilities unconfident. During this stage, we want the target feature representations to be clustered tightly around the target prototypes (confident network outputs). % \mz{need more elaboration} \kt{updated} 
The target refinement loss is thus formulated as:
$\mathcal{L}_{target\_refinement}(G^t; \mathcal{X}^t) = \mathcal{L}_{transport}(G^t; \mathcal{X}^t) + \mathcal{L}_{mi}(G^t; \mathcal{X}^t).\label{eq:target_refinement}$

%After we obtain a well-initialized target model, we follow the same procedure as the source training to learn the target model.

\vspace{-2mm}

\section{Related work}
\vspace{-2mm}

\BF{Clustering.}
 For a complete picture of the field, readers may refer to the survey by \citet{min2018survey}. We emphasize deep-clustering-based approaches, which attempt to learn the feature representation of the data while simultaneously discovering the underlying clusters: K-means \citep{yang2017towards, caron2018deep}, information maximization \citep{menapace2020learning, ji2019invariant, kim2021contrastive, do2021clustering}, transport alignment\citep{asano2019self, caron2020unsupervised, wang2022representing}, neighborhood-clustering \citep{xie2016unsupervised, huang2019unsupervised, dang2021nearest}, contrastive learning \citep{pan2021multi, shen2021you}, probabilistic approaches \citep{ yang2020adversarial, monnier2020deep, falck2021multi, manduchi2021deep}, and kernel density \citep{yang2021discriminative}.
These works primarily focus on clustering data for downstream tasks for a single domain, whereas our clustering algorithm is designed to cluster the data from multiple domains. Moreover, our method solves the problem of transferring the knowledge from the data-rich source domain to the target domain. Distinct from ACIDS \citep{menapace2020learning} which maximizes the mutual information between different views of the same image, our method maximizes the mutual information between cluster labels and images. In addition to data privacy, we also consider model privacy.

\BF{Source-free knowledge transfer.}
Early domain adaptation methods \citep{ ben2006analysis, blitzer2006domain, tzeng2014deep, ganin2015dann, long2017jan, long2018cdan,long2015dan,tzeng2017adversarial, courty2017jdot} focus on reducing the distributional discrepancies between the source and target domain data. These methods, however, require access to the source and target data simultaneously during the adaptation process, compromising the privacy of the source domain. To overcome this issue, several methods \citep{kuzborskij2013stability, du2017hypothesis, liang2020we, li2020model, kundu2020universal, kurmi2021domain, yeh2021sofa, tanwisuth2021prototype} have been developed for source data-free domain adaptation. For a more thorough literature review of this field, we refer the reader to the survey paper by \citet{yang2021generalized}. In contrast to those methods, we consider a more challenging adaptation setting, as used in previous works \citep{lipton2018detecting, deng2021universal,liang2021distill, zhang2021unsupervised}, where the privacy of both data and models is the main concern. Different from these lines of work, our approach relies on labeled data in neither the source nor target domain.
 
\vspace{-2mm}

\section{Experiments}
\vspace{-3mm}

\label{sec:experiments}
In this section, we evaluate our method on Office-31, Office-Home, and PACS datasets under three different transfer learning scenarios. The first setting (standard setting) includes only input distribution shift. The second setup (model transfer setting) contains both input and model shifts. The last scenario (limited-data and cluster-imbalanced setting) involves both input and cluster-proportion shifts.
\vspace{-2mm}

\subsection{Experimental setup}
\vspace{-2mm}

\BF{Comparable methods.} 
We benchmark against existing clustering approaches---{\it DeepCluster of \citet{caron2018deep}, Invariant Information Clustering (IIC) of \citet{ji2019invariant}}, and {\it Adaptive Clustering of Images under Domain Shift  (ACIDS) of \citet{menapace2020learning}}---in the UCDS setting when the results are available. Unless specified otherwise,  the reported baseline results are directly taken from \citet{menapace2020learning}. IIC and DeepCluster train on target data only while ACIDS trains a source model and then adapts on the target data. We also compare our approach to the following alternative methods, which are different components of our framework: {\it Pre-trained Only (PO)}, which uses a pre-trained network to cluster target data directly; {\it Source training Only (SO)}, which trains a model on all the source data using Eq. \eqref{eq:source_joint_loss} and directly tests on the target data; {\it Target Training Only (TO)}, which trains a model on the target data using  the loss in Section \ref{eq:target_refinement} without source knowledge transfer; {\it Adaptation Only (AO)}, which performs the first two stages of our framework, source model training and target model clustering, without further refining on the target data; {\it PCD (Ours)} refers to using all three stages of our approach: source model learning, target model clustering, and target model refinement. 
SO allows us to see the significance of the source model training. Compared with PCD, TO enables us to evaluate the importance of the source knowledge transfer, while AO helps us see the improvement from target refinement.  % Describe deep clustering and ACIDs a bit
% (4) \textsc{Mutual Information} \sz{MI}, which trains a trains a model on the target data only using the mutual-information objective in Eq \ref{}. 

\BF{Pre-trained networks.} To verify the compatibility of our approach with different models, we consider multiple types of pre-trained network architectures and pre-training schemes in our experiments. For pre-training schemes, we explore supervised and self-supervised pre-trainings on ImageNet \citep{russakovsky2015imagenet}. For network architectures, we experiment with supervised ResNet-18 as well as self-supervised ResNet-50 \citep{he2016deep} and Vision Transformer (ViT) \citep{dosovitskiy2020image}. In particular, we adopt the network trained by SWAV \citep{caron2020unsupervised} for ResNet-50 and that trained by DINO \citep{caron2021emerging} for Vision Transformer for our self-supervised pre-training.

\BF{Datasets and evaluation metric.}
We use the following datasets in our experiments: Office-31 \citep{saenko2010adapting}, Office-Home \citep{venkateswara2017deep}, and PACS \citep{li2017deeper}. The Office-31 dataset has three domains (Amazon, Webcam, DSLR) with 4,652 images. The Office-Home dataset consists of 15,500 images with four domains (Art, Clipart, Product, and Real-world). The PACS dataset contains four domains (Art, Painting, Cartoon, and Sketch) with 9,991 images. Following prior works \citep{ji2019invariant,menapace2020learning}, we evaluate all methods using clustering accuracy on the target dataset. The metric is calculated by first solving a linear assignment problem to match the clusters to ground-truth classes. We set $K$, the number of clusters, equal to the number of classes in each dataset for evaluation purposes.

\BF{Implementation details.}
We follow the standard protocols for source-free domain adaptation \citep{liang2020we}. Specifically, we use mini-batch SGD with a momentum of $0.9$ and weight decay of $0.001$. Both source and target encoders are initialized with ImageNet pre-trained networks \citep{russakovsky2015imagenet}, but the prototypes and the projection layer of the encoder are initialized with a random linear layer. The initial learning rates are set to $0.001$ for the pre-trained encoders and $0.01$ for the randomly initialized layer. The learning rates, $\eta$, follows the following schedule: $\eta=\eta_0(1+10p)^{-0.75}$ where $\eta_0$ is the initial learning rate. We use the batch size of $64$ in both source and target learning. All three loss terms are equally weighted, while other choices are possible. We report the sensitivity of the coefficients in front of the loss terms in Appendix \ref{appendix:sensitivity}. The initial value of $\beta_0$ to learn domain-specific proportions is set to $0.9999$ for source clustering and $0.99$ for target clustering in all settings. We run our method with three different random seeds to calculate the standard deviation. Full implementation details are included in Appendix \ref{appendix:implementation} 

% The source model and hyper-parameters are selected using the validation set of the source domain. The target model is trained using all the target data. 
%\sz{remove or move above sentence}. \sz{also shorten this paragraph and move some sentences to the appendix}
 \vspace{-1mm}
\begin{table}[htp!]
\centering
 \caption{Clustering accuracy $(\%)$ on different datasets for ResNet-18-based methods. $\mathcal{R}$ denotes the rest of the domains.  } 
 \vspace{-2mm}
 \resizebox{1.0\textwidth}{!}{\begin{tabular}{c|ccc|c|cccc|c|cccc|c}
\toprule  Settings & \multicolumn{4}{c|}{Office-31} & \multicolumn{5}{c|}{Office-Home} & \multicolumn{5}{c}{PACS}  \\
\cmidrule(r){2-5} \cmidrule(r){6-10} \cmidrule{11-15} 
\multicolumn{1}{c}{} & \multicolumn{1}{|c}{$\mathcal{R} \rightarrow$ A} & \multicolumn{1}{c}{$\mathcal{R} \rightarrow$ W} & \multicolumn{1}{c}{$\mathcal{R} \rightarrow$ D} & \multicolumn{1}{|c|}{Avg} & \multicolumn{1}{c}{$\mathcal{R} \rightarrow$ Ar} & \multicolumn{1}{c}{$\mathcal{R} \rightarrow$ Cl} & \multicolumn{1}{c}{$\mathcal{R} \rightarrow$ Pr} & \multicolumn{1}{c}{$\mathcal{R} \rightarrow$ Rw} & \multicolumn{1}{|c|}{Avg} & \multicolumn{1}{c}{$\mathcal{R} \rightarrow$ P} & \multicolumn{1}{c}{$\mathcal{R} \rightarrow$ A}& \multicolumn{1}{c}{$\mathcal{R} \rightarrow$ C}& \multicolumn{1}{c}{$\mathcal{R} \rightarrow$ S}& \multicolumn{1}{|c}{Avg}\\
\midrule
DeepCluster \citep{caron2018deep} & $19.6$ & $18.9$  & $18.7$  & $19.1$ & $8.9$ & $11.1$ & $16.9$ & $13.3$ & $12.6$ & $27.9$  & $22.2$ & $24.4$ & $27.1$ & $25.4$\\
IIC \citep{ji2019invariant} & $31.9$ & $37.0$  & $34.0$  & $34.4$ & $12.0$ & $15.2$ & $22.5$ & $15.9$ & $16.4$ & $70.6$ &  $39.8$ & $39.6$ & $46.6$ & $49.2$\\
ACIDS \citep{menapace2020learning} & $33.4$ & $37.5$  & $36.1$ & $35.7$ & $12.0$  & $16.2$ & $23.9$ & $15.7$ & $17.0$ & $64.4$ &  $42.1$ & 44.5 & $51.1$  & $50.5$\\
\midrule

PO & $14.1$ &  $17.9$ & $18.3$ & $16.8$ & $11.4$ & $9.0$ & $12.9$ & $10.8$ & $11.0$ & $30.5$  & $24.1$ & $19.8$ & $20.8$ & $23.8$ \\
SO & $34.5$ & $46.7$ & $43.0$ & $41.4$ & $23.6$ & $15.6$ & $23.1$ & $21.8$ & $21.0$ & $30.8$  & $35.7$ & $27.6$ & $26.0$ & $30.0$\\
TO & $38.0$ & $46.6$  & $45.3$  & $43.3$ & $21.3$ & $12.2$ & $30.6$ & $24.2$ & $22.1$ & $88.4$ & $\mathbf{56.5}$ & $56.5$ & $49.1$ & $62.6$ \\
AO & $42.8$ & $58.4$  & $55.8$  & $52.3$ & $30.0$ & $22.7$ & $29.3$ & $24.4$ & $26.6$ & $91.5$ & $47.7$ & $52.3$ & $49.1$ & $60.2$\\
\midrule

 $\mathbf{PCD}$ & $\mathbf{46.8}$ & $\mathbf{60.0}$ & $\mathbf{57.8}$ & $\mathbf{54.9}$  & $\mathbf{33.3}$ & $\mathbf{24.4}$ & $\mathbf{31.4}$ & $\mathbf{28.1}$ & $\mathbf{29.3}$ & $\mathbf{92.6}$ & 49.7 & $\mathbf{56.7}$ & $\mathbf{53.4}$ & $\mathbf{63.4}$\\
\bottomrule
\end{tabular}}
\label{tab:resnet18}
\end{table}
\vspace{-2mm}

\subsection{Main results}
\vspace{-2mm}

\BF{Standard setting.} In real-world applications, the source and target data distributions often differ. To test our method under input distribution shift, we evaluate our method on Office-31, Office-Home, and PACS datasets. For each experiment, we select one domain as the target and all the other, % rest of the domains, 
denoted as $\mathcal{R}$, as the source domains. We use the same model architecture in both the source and target domains. We report the results for ResNet-18 (supervised pre-training) in Table \ref{tab:resnet18}. The full results with standard error are shown in Appendix \ref{appendix:full_results}. Compared with the results reported by \citet{menapace2020learning}, our algorithm outperforms ACIDS consistently in all three datasets (see Table \ref{tab:resnet18}): $19.2\%$ on Office-31, $12.3\%$ on Office-Home, and $12.9\%$ on PACS. Though ACIDS does not address the problem of our setting with the same pre-training scheme and backbones as our method, we report the results for comparison. The results of ACIDS with this pre-training scheme are included in Appendix~\ref{appendix:full_results} in Table~\ref{tab:acids_comparison}. We observe that our approach still outperforms ACIDS on three out of four tasks with 4\% higher in the average accuracy, emphasizing the general applicability and strong performance of PCD. With no adaptation, TO achieves higher clustering accuracy than both IIC and DeepCluster, demonstrating the effectiveness of our clustering method.

Compared with our own alternative methods ($i.e.,$ PO, SO, TO, and AO), PCD achieves steady gains in performance except for one task. Notably, on the task $\mathcal{R} \rightarrow$ A of the PACS dataset, we notice a negative transfer \citep{wang2019characterizing} as TO performs the best ($56.5\%$ vs. $49.7\%$). We hypothesize that the Art domain looks quite distinct from the source domain data, and the supervised-pretraining backbone is strong enough to yield good performance using target training only. SO improves upon PO on all the tasks, showing that the knowledge from the source domain can benefit the target domain learning. Likewise, we see consistent improvements around $2-3\%$ over AO . This result illustrates the importance of target model refinement. We observe similar patterns using self-supervised ResNet-50 as the backbone (see Appendix \ref{appendix:full_results}).

\BF{Model transfer setting.}  
In many applications, source and target data owners may have different resource requirements. As an example, unlike source providers such as Google, target clients may have limited resources. Thus, they may not be able to use the same model architecture as the source provider. To illustrate the flexibility and demonstrate the generalizability of our framework under model shift, we experiment with different model architectures and pre-training schemes in the source and target domains. We explore three different combinations of source and target model architectures and pre-training schemes: ViT-B/16 (self-supervised) $\rightarrow$ ResNet-50 (self-supervised), ViT-B/16 (self-supervised) $\rightarrow$ ResNet-50 (supervised), and ViT-B/16 (self-supervised) $\rightarrow$ ResNet-18 (supervised). The results are reported in Table \ref{tab:model_trans}. In both settings, we continually see improvements in average performance. This finding shows that our method still performs well even though the source and target domain architectures differ, providing strong evidence for the generalizability and compatibility of different components of our framework.

%This finding shows that our method still outperform the baselines when the architectures of the source and target domain differ, demonstrating the generalizability of our method.
%Table shows that consistently improves upon the corresponding clustering 
%models on not only same, which confirms our results in Section , but also n. The performance gains on are often greater than the gains on , meaning that can significantly help the model to generalize across domains

%In the label-free pipeline ViT-B/16 $\rightarrow$ Resnet-50, we observe around $3\%$ improvement over Resnet-50 $\rightarrow$ Resnet-50, showing the benefit of transferring from a larger model in the source domain. 

%However, ViT-B/16 (self-supervised) $\rightarrow$ Resnet-18 (supervised) beats Resnet-18 (supervised) $\rightarrow$ Resnet-18 (supervised) in one out of three tasks. We hypothesize that the different pre-training schemes are not compatible with each other.

%The gap in the average clustering accuracy between TO and our method is wider ($14.9\%$ vs. $8.3\%$) when we use self-supervised pre-training as the backbone, underlining the benefits of our method in the label-free pipeline. 

\begin{table}[t!]
%\vspace{-4mm}
\centering
 \caption{Clustering accuracy $(\%)$ on Office-31 for different model transfer settings. {\it ssl} and {\it sup} denote self-supervised and supervised pre-trainings, respectively. (source) $\rightarrow$ (target).} 
 \vspace{-2mm}\resizebox{1.0\textwidth}{!}{\begin{tabular}{c|ccc|c|ccc|c|ccc|c}
\toprule  Settings & \multicolumn{4}{c|}{ViT-B/16 (ssl) $\rightarrow$ ResNet-50 (ssl)} & \multicolumn{4}{c|}{ViT-B/16 (ssl) $\rightarrow$ ResNet-50 (sup)} & \multicolumn{4}{c}{ViT-B/16 (ssl) $\rightarrow$ ResNet-18 (sup)}   \\
\cmidrule(r){2-5} \cmidrule(r){6-9} \cmidrule(r){10-13} 
\multicolumn{1}{c|}{} & \multicolumn{1}{c}{$\mathcal{R} \rightarrow$ A} & \multicolumn{1}{c}{$\mathcal{R} \rightarrow$ W} & \multicolumn{1}{c}{$\mathcal{R} \rightarrow$ D} & 
\multicolumn{1}{|c}{Avg} & \multicolumn{1}{|c}{$\mathcal{R} \rightarrow$ A} & \multicolumn{1}{c}{$\mathcal{R} \rightarrow$ W} & \multicolumn{1}{c}{$\mathcal{R} \rightarrow$ D} & 
\multicolumn{1}{|c}{Avg}&
\multicolumn{1}{|c}{$\mathcal{R} \rightarrow$ A} & \multicolumn{1}{c}{$\mathcal{R} \rightarrow$ W} & \multicolumn{1}{c}{$\mathcal{R} \rightarrow$ D} & 
\multicolumn{1}{|c}{Avg} \\
\midrule
PO &  $20.1/13.5$ &  $26.7/16.7$ & $27.2/19.3$ & $24.7/16.5$ & $20.1/15.7$ &  $26.7/24.2$ & $27.2/18.8$ & $24.7/19.6$ & $20.1/14.1$ & $26.7/17.9$ & $27.2/18.3 $ & $24.7/16.8$ \\
SO  & $43.2$ & $46.4$ & $37.2$ & $42.3$ & $43.2$ & $46.4$ & $37.2$ & $42.3$ & $43.2$ & $46.4$ & $37.2$ & $42.3$\\
TO  & $32.6$ & $34.3$ & $33.7$ & $33.5$ & $43.7$ & $55.8$ & $\mathbf{52.0}$ & $50.5$ & $38.0$ & $45.3$ & $46.6$ & $43.3$ \\
AO & $50.6$ & $49.7$ & $36.4$ & $45.6$ & $52.5$ & $53.7$ & $44.2$ & $50.1$ & $53.0$ & $47.2$ & $43.9$ & $48.0$ \\
\midrule
$\mathbf{PCD}$ & $\mathbf{51.7}$ &  $\mathbf{51.7}$ & $\mathbf{41.8}$ & $\mathbf{48.4}$ &  $\mathbf{54.4}$ & $\mathbf{60.8}$ & $49.2$ & $\mathbf{54.8}$  &$\mathbf{54.6}$ & $\mathbf{53.6}$ & $\mathbf{46.7}$ & $\mathbf{51.6}$ \\
\bottomrule
\end{tabular}}
\label{tab:model_trans}
\vspace{-3mm}
\end{table}

\BF{Limited-data and cluster-imbalanced setting.} 
In real-world scenarios, target domain data are often scarce and imbalanced. To further show the benefit of our clustering loss under this setting, we follow the experimental procedures in \citet{tachet2020labelshift} and \citet{zhang2022allsh}. Specifically, we drop 70\% of the target data in the first $\left\lfloor K/2 \right\rfloor$ clusters to create this scenario. The experiments are done on the Office-31 dataset. To illustrate the use of our method in a label-free pipeline, we utilize self-supervised ResNet-50 as the feature encoder for both source and target domains. This scenario is extremely challenging for transfer learning methods since there are shifts in both image and cluster-label distributions. However, as we see in Table \ref{tab:limited_imbalanced}, PCD still outperforms TO by $4\%$. We note that TO also adaptively learns the target proportions but does not have to deal with distribution shifts. We also observe consistent improvements over other alternative methods. This result highlights the use of our method in practical settings with limited and imbalanced data.

%This result is not surprising since we incorporate proportion estimation in both the source and target domains.
%Even though the limited and imbalanced data in the target domain make clustering more challenging, the performance does not drop significantly from that in the regular setting (from 45.6 to 43.7). 

\begin{table}[htp!]
\centering
 \caption{Clustering accuracy $(\%)$ on sub-sampled version of Office-31 for ResNet-50-based methods.}
 \vspace{-2mm}
 \resizebox{0.45\textwidth}{!}{\begin{tabular}{c|cccc}
\toprule  Settings & $\mathcal{R} \rightarrow$ sub-A & $\mathcal{R} \rightarrow$ sub-W & $\mathcal{R} \rightarrow$ sub-D  &   Avg \\
\midrule
PO & $14.6$ & $16.7$ & $21.5$ & $17.6$\\
SO & $21.1$ & $32.5$ & $36.0$ &  $29.9$\\
TO & $31.4$ & $41.9$ & $45.1$ & $39.5$\\
AO& $34.7$ & $40.8$ & $43.9$ & $39.8$\\
\midrule
$\mathbf{PCD}$ & $\mathbf{37.8} $ &	$\mathbf{46.4}$	& $\mathbf{47.0}$	& 	$\mathbf{43.7}$\\
\bottomrule
\end{tabular}}
\vspace{-4mm}
\label{tab:limited_imbalanced}
\end{table}

% \begin{table}[htp!]
% \vspace{-4mm}
% \centering
%  \caption{Clustering accuracy $(\%)$ on different versions of Office-31 for Resnet-50-based methods.} \resizebox{0.8\textwidth}{!}{\begin{tabular}{c|ccc|c|ccc|c}
% \toprule  Settings & \multicolumn{4}{c}{Regular setting} & \multicolumn{4}{|c}{Limited and imbalaced setting}\\
% \cmidrule(r){2-5} \cmidrule(r){6-9}  
%  \multicolumn{1}{c|}{}& \multicolumn{1}{c}{$\mathcal{R} \rightarrow$ A} & \multicolumn{1}{c}{$\mathcal{R} \rightarrow$ W} & \multicolumn{1}{c}{$\mathcal{R} \rightarrow$ D}  &   \multicolumn{1}{|c|}{Avg}  & \multicolumn{1}{c}{$\mathcal{R} \rightarrow$ sub-A} & \multicolumn{1}{c}{$\mathcal{R} \rightarrow$ sub-W} & \multicolumn{1}{c}{$\mathcal{R} \rightarrow$ sub-D}  &   \multicolumn{1}{|c}{Avg} \\
% \midrule
% PO & $13.5$ &  $16.7$ & $19.3$ & $16.5$ & $14.6$ & $16.7$ & $21.5$ & $17.6$\\
% SO & $19.0$ & $26.3$ & $27.5$ & $24.3$ & $21.1$ & $32.5$ & $36.0$ &  $29.9$\\
% TO & $31.6$ & $34.3$  & $33.7$  & $33.2$  & $31.4$ & $41.9$ & $45.1$ & $39.5$\\
% AO& $33.3$ & $37.6$  & $41.9$  & $37.6$ & $34.7$ & $40.8$ & $43.9$ & $39.8$\\
% \midrule
% $\mathbf{PCD}$ & $\mathbf{37.8}$ & $\mathbf{48.2}$ & $\mathbf{51.0}$	& $\mathbf{45.6}$ & $\mathbf{37.8}$ &	$\mathbf{46.4}$	& $\mathbf{47.0}$	& 	$\mathbf{43.7}$\\
% \bottomrule
% \end{tabular}}
% \label{tab:limited_imbalanced}
% \end{table}

\begin{figure}[t!]
    \centering
        \includegraphics[width=1.0\textwidth]{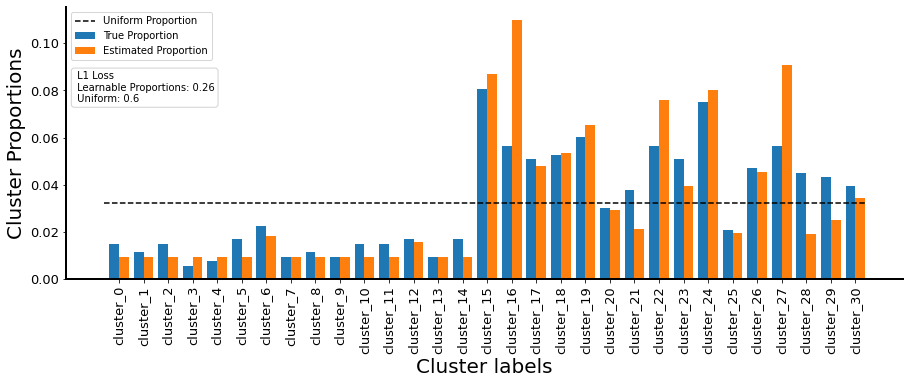}%
        %\vspace{-1mm}
        \caption{Visualization of the cluster proportions for the sub-sampled version of the task $\mathcal{R}$ $\rightarrow$ sub-W on the Office-31 dataset. To create this plot, we first match the predicted clusters with the true labels using optimal assignment. The blue bars exhibit the true cluster proportions, whereas the orange bars depict the estimated cluster proportions. The L1 loss of the estimated cluster distribution is lower than that of the uniform proportion (0.26 vs. 0.6), demonstrating the success of our estimation. 
       }
        \vspace{-2mm}
        \label{fig:proportions}%
\end{figure}

\begin{figure}
[t!] %htp!]

\vspace{-6mm}

\centering
  \subfloat[]{%
        \includegraphics[width=0.33\textwidth]{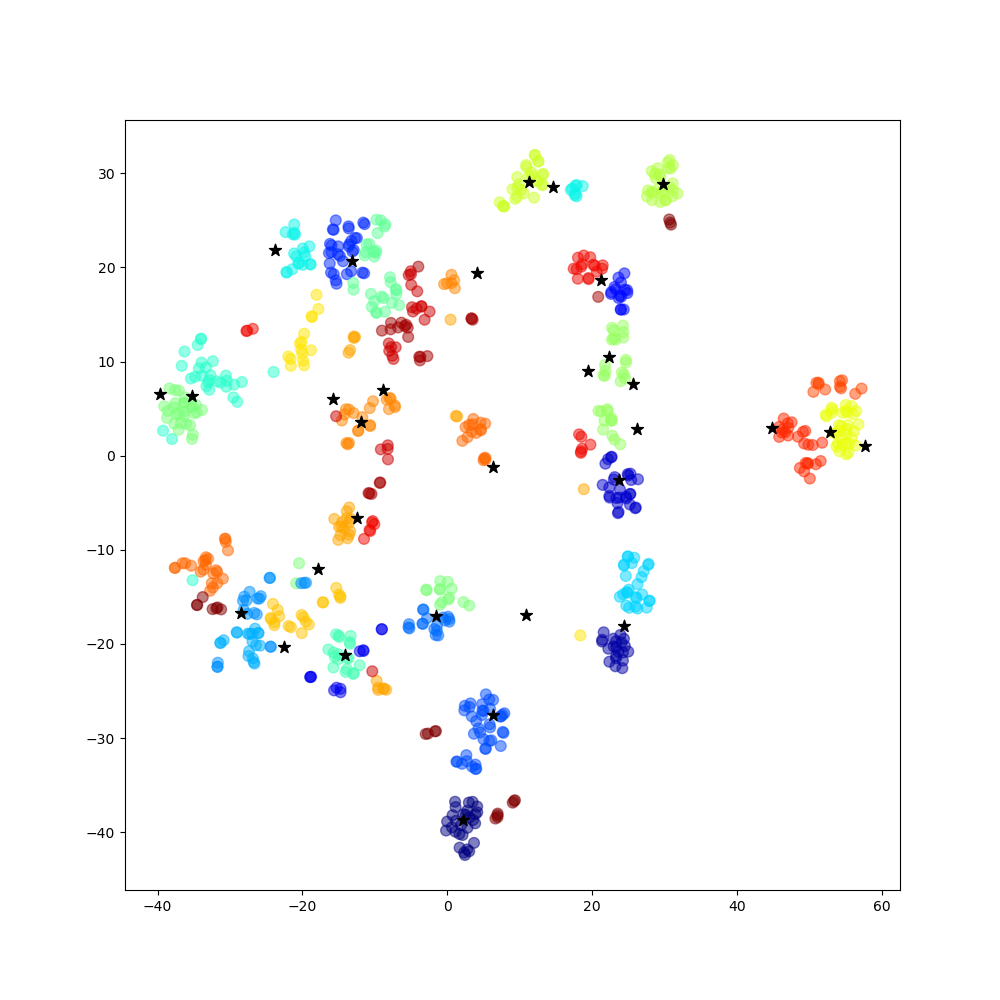}%
        \label{fig:tsne_1}%
        }%
    \subfloat[]{%
        \includegraphics[width=0.33\textwidth]{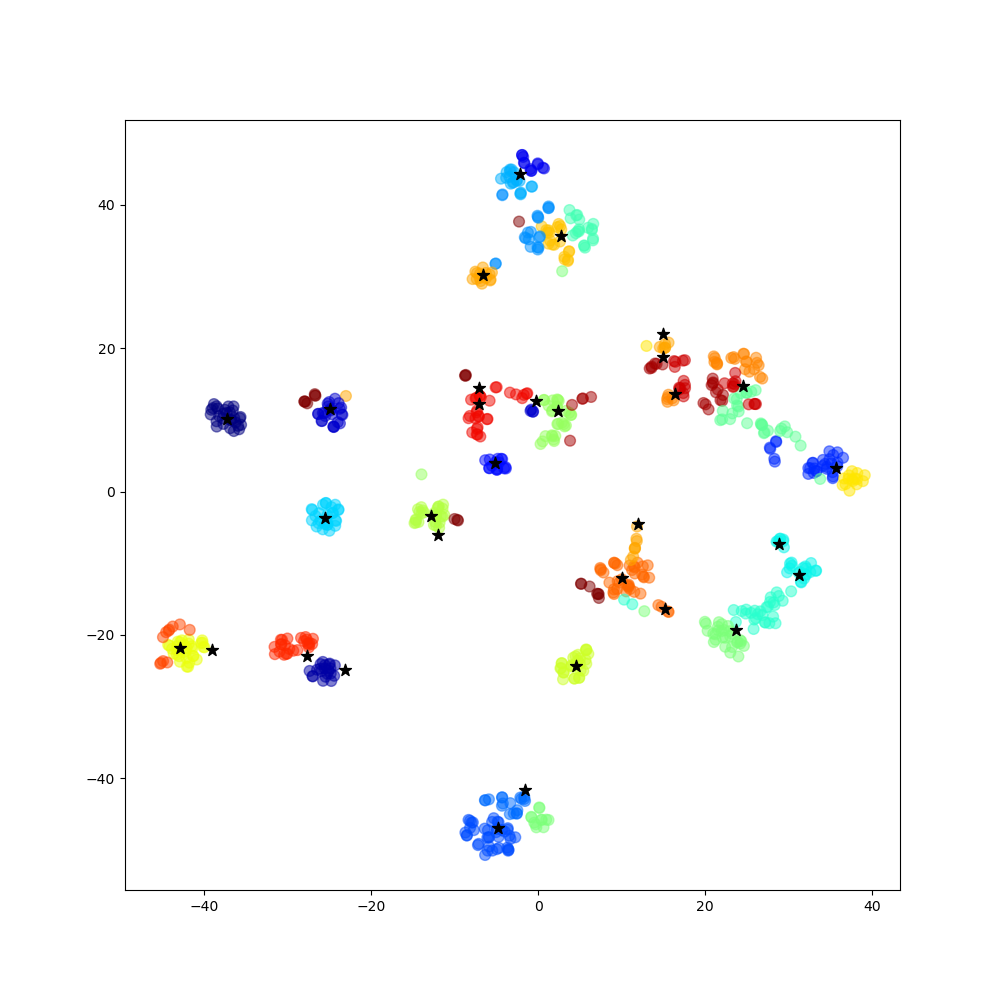}%
        \label{fig:tsne_2}%
        }%
    % \subfloat[]{%
    %     \includegraphics[width=0.25\textwidth]{figures/A2W_tsne_.PNG}%
    %     \label{fig:tsne_domain}%
    %     }%
    \vspace{-2mm}
     \caption{t-SNE visualizations of the encoder's outputs on the task $\mathcal{R}\rightarrow$ W. Different colors represent semantic classes from the ground-truth labels. Figure (a) shows the outputs trained with Target training Only (TO), while Figure (b) depicts those trained with the whole framework. Samples with similar semantic content are more tightly clustered around the prototypes ($\star$) in Figure (b).} \vspace{-2mm}
\end{figure}
\vspace{-2mm}
\section{Analysis}
\vspace{-2mm}

\BF{Ablation study.}
To see the contribution of each component, we remove one part at a time from the whole framework and present the results in Table \ref{tab:ablation}. 
Overall, PCD achieves higher clustering accuracy than all other alternative versions with privacy constraints. We observe that the clustering accuracy drops dramatically ($10.3\%$) without the prototype clustering, illustrating the importance of this element. The mutual-information objective is also significant since omitting it leads to a drop in clustering accuracy of $7.3\%$. This observation shows that the two losses are complementary to each other. The temporal ensemble of the cluster labels produced by the source model still improves the model but does not significantly hurt the performance if removed. We also report the result of directly initializing the target model with the source model (w/o model privacy). We notice around $3\%$ improvement. When pooling all the source domains together into a single source domain for clustering (pooled source), we see a drop of $2.2\%$. This result indicates that we should respect the local structures of data in each domain.

\begin{table}[!htp]
\centering
\caption{Clustering accuracy (\%) on the task $\mathcal{R}$ $\rightarrow$ W (Office-31) under different variants  (ResNet-18).}
\vspace{-2mm}
 \resizebox{0.8\columnwidth}{!}{ \begin{tabular}{ccccccc}
\toprule
Full & w/o prototype clustering &
 w/o MI &  w/o CutMix &  w/o Temporal Ensemble & w/o model privacy & pooled source \\
\midrule
 $60.0 $ & $49.7$& $52.7 $ & $54.5 $ & $58.3$ & $63.1$ & $57.8$ \\
\bottomrule
\end{tabular}}
\label{tab:ablation}
\vspace{-3mm}
\end{table}
%\mz{where is the ablation study that aggregates all the sources domains into a single source domain?}

\BF{Results analysis.} {\it Visualization.}
In Figure \ref{fig:proportions}, we visualize the estimated target proportions versus the true proportions, which are calculated from the ground-truth labels. The learned cluster distribution achieves lower L1 loss than the uniform distribution, meaning that the estimated values reflect the data distribution better than the uniform proportions. We plot the t-SNE visualization of the outputs of the feature encoder for the model trained with target training only (Section \ref{eq:target_refinement}) in Figure \ref{fig:tsne_1} and the one trained with the whole framework (Algorithm \ref{alg:pseudo_code}) in Figure %\ref{fig:tsne_1} and 
\ref{fig:tsne_2}. Using the whole framework, we can see that the samples are more tightly clustered around the prototypes, illustrating that the knowledge from the source domain benefits the target model learning.
{\it Running time and parameter size.} We report the number of parameters and running time per step for comparison in Appendix \ref{appendix:param_size}, where we see that our method is more efficient in both time and memory than ACIDS.

% TODO: plot the proportions in the target domain.
% TODO: Separate vs joint training in the target.
% TODO: Study on SeLa.
% TODO: Hyper-parameter sensitivity plots.
% TODO: Study on selecting $K$ based on the loss.
% TODO: implementation details, limited data, justification for K
% TODO: compare our results with existing baselines in the experiment section
% TODO: hyper-parameters description
% TODO: describe how model initialization 
% TODO: too many loss functions. present it better. too many begin align
% TODO: add description in the main text and hyper-parameter plots in the appendix.

% TODO Experiments: consistent gains, confirm conjecture, provide justification for some performance
\vspace{-2mm}
\section{Conclusion}
\vspace{-2mm}
We study a practical transfer learning setting that does not rely on labels in the source and target domains and considers the privacy of both the source data and model. To solve this problem, we provide a novel solution that utilizes prototype clustering, mutual-information maximization, data augmentation, and knowledge distillation. Experiments show that our clustering approach consistently outperforms the baselines and works well across different datasets and model architectures.

%we show through experiments (TODO) that, in the representation learning step, using the cosine loss instead of cross-entropy can yield better performance in our setting. Finally,
\newpage
%%%%%%%%%%%%%%%%%%%%%%%%%%%%%%%%%%%%%%%%%%%%%%%%%%%%%%%%%%%%
\bibliographystyle{unsrtnat}
\bibliography{reference.bib}
%%%%%%%%%%%%%%%%%%%%%%%%%%%%%%%%%%%%%%%%%%%%%%%%%%%%%%%%%%%%

\newpage
\appendix
\begin{center}
    {\textbf{\LARGE A Prototype-oriented Clustering for Domain Shift \\
    \vspace{2mm} with Source Privacy: Appendix}}\\
    \vspace{4mm}
\end{center}

% \section{Broader impact}
% \label{sec:broader_impact}
% Any approaches that deal with clustering can suffer from domain shift which is caused by dataset bias. While adaptation methods can help alleviate the domain shift between different domains, it cannot eliminate the issue because of the combinatorial nature of too many exogenous factors \citep{gong2012overcoming}. As with any computationally intensive venture, attention must be given to the use of sustainable energy sources. On the bright side, our method can help protect the privacy of both data and model while helping with domain shift.

 %While our method can protect the privacy of the source data and model, it is still subject to different model attacks to replicate the model or recover the data distribution \citep{kariyappa2021maze, miura2021megex}. We leave it for future work to address this limitation.
 
\section{Full experimental results}
\label{appendix:full_results}
\subsection{Standard setting}

\begin{table}[htp!]
\vspace{-4mm}
\centering
 \caption{Clustering accuracy $(\%)$ on different datasets for ResNet-18-based methods (supervised pre-training for all methods below the mid line) and (random initialization for all methods above the mid line). } \resizebox{1.0\textwidth}{!}{\begin{tabular}{c|ccc|c|cccc|c|cccc|c}
\toprule  Settings & \multicolumn{4}{c|}{Office-31} & \multicolumn{5}{c|}{Office-Home} & \multicolumn{5}{c}{PACS}  \\
\cmidrule(r){2-5} \cmidrule(r){6-10} \cmidrule{11-15} 
\multicolumn{1}{c}{} & \multicolumn{1}{|c}{$\mathcal{R} \rightarrow$ A} & \multicolumn{1}{c}{$\mathcal{R} \rightarrow$ W} & \multicolumn{1}{c}{$\mathcal{R} \rightarrow$ D} & \multicolumn{1}{|c|}{Avg} & \multicolumn{1}{c}{$\mathcal{R} \rightarrow$ Ar} & \multicolumn{1}{c}{$\mathcal{R} \rightarrow$ Cl} & \multicolumn{1}{c}{$\mathcal{R} \rightarrow$ Pr} & \multicolumn{1}{c}{$\mathcal{R} \rightarrow$ Rw} & \multicolumn{1}{|c|}{Avg} & \multicolumn{1}{c}{$\mathcal{R} \rightarrow$ P} & \multicolumn{1}{c}{$\mathcal{R} \rightarrow$ A}& \multicolumn{1}{c}{$\mathcal{R} \rightarrow$ C}& \multicolumn{1}{c}{$\mathcal{R} \rightarrow$ S}& \multicolumn{1}{|c}{Avg}\\
\midrule

DeepCluster & $19.6$ & $18.9$  & $18.7$  & $19.1$ & $8.9$ & $11.1$ & $16.9$ & $13.3$ & $12.6$ & $27.9$  & $22.2$ & $24.4$ & $27.1$ & $25.4$\\
IIC & $31.9$ & $37.0$  & $34.0$  & $34.4$ & $12.0$ & $15.2$ & $22.5$ & $15.9$ & $16.4$ & $70.6$ &  $39.8$ & $39.6$ & $46.6$ & $49.2$\\
ACIDS & $33.4$ & $37.5$  & $36.1$ & $35.7$ & $12.0$  & $16.2$ & $23.9$ & $15.7$ & $17.0$ & $64.4$ &  $42.1$ & 44.5 & $51.1$  & $50.5$\\
\midrule
PO & $14.1\pm{1.6}$ &  $17.9\pm{2.0}$ & $18.3\pm{2.9}$ & $16.8$ & $11.4\pm{1.6}$ & $9.0\pm{1.6}$ & $12.9\pm{2.8}$ & $10.8\pm{1.7}$ & $11.0$ & $30.5\pm{3.1}$  & $24.1\pm{0.6}$ & $19.8\pm{3.7}$ & $20.8\pm{1.7}$ & $23.8$ \\
SO & $34.5\pm{0.5}$ & $46.7\pm{2.9}$ & $43.0\pm{2.9}$ & $41.4$ & $23.6\pm{1.6}$ & $15.6\pm{1.9}$ & $23.1\pm{3.7}$ & $21.8\pm{2.9}$ & $21.0$ & $30.8\pm{8.2}$  & $35.7\pm{3.9}$ & $27.6\pm{8.3}$ & $26.0\pm{3.7}$ & $30.0$\\
TO & $38.0 \pm{3.2}$ & $46.6 \pm{1.6}$ & $45.3 \pm{1.5}$  & $43.3$ & $21.3 \pm{2.6}$ & $12.2 \pm{0.7}$& $30.6\pm{4.1}$ & $24.2\pm{0.7}$ & $22.1$ & $88.4\pm{3.9}$  & $\mathbf{56.5}\pm{4.1}$ & $56.5\pm{11.1}$ & $49.1\pm{2.8}$ & $62.6$ \\
AO & $42.8\pm{0.9}$ & $58.4\pm{3.7}$  & $55.8\pm{1.9}$  & $52.3$ & $30.0\pm{1.7}$ & $22.7\pm{1.6}$ & $29.3\pm{4.1}$ & $24.4\pm{2.6}$ & $26.6$ & $91.5\pm{5.9}$ & $47.7\pm{5.7}$ & $52.3\pm{1.2}$ & $49.1\pm{3.0}$ & $60.2$\\
\midrule
$\mathbf{PCD}$ & $\mathbf{46.8} \pm{1.7}$ & $\mathbf{60.0} \pm{2.6}$ & $\mathbf{57.8} \pm{5.9}$ & $\mathbf{54.9} $  & $\mathbf{33.3} \pm{1.0}$ & $\mathbf{24.4} \pm{1.5}$ & $\mathbf{31.4} \pm{4.7}$ & $\mathbf{28.1} \pm{2.5}$ & $\mathbf{29.3}  $ & $\mathbf{92.6} \pm{2.4}$ & $49.7\pm{5.0}$ & $\mathbf{56.7} \pm{2.6}$ & $\mathbf{53.4}\pm{5.9}$ & $\mathbf{63.4}$\\
\bottomrule
\end{tabular}}
\label{tab:resnet18_comparison}
\end{table}

\begin{table}[htp!]
\vspace{-4mm}
\centering
 \caption{Clustering accuracy $(\%)$ on different datasets for ResNet-18-based methods (supervised pre-training).} \resizebox{0.5\textwidth}{!}{
 \begin{tabular}{c|cccc|c|}
\toprule  Settings & \multicolumn{5}{c|}{PACS}  \\
\cmidrule(r){2-6} 
\multicolumn{1}{c}{} & \multicolumn{1}{|c}{$\mathcal{R} \rightarrow$ P} & \multicolumn{1}{c}{$\mathcal{R} \rightarrow$ A} & \multicolumn{1}{c}{$\mathcal{R} \rightarrow$ C} &
\multicolumn{1}{c}{$\mathcal{R} \rightarrow$ S} &
\multicolumn{1}{|c|}{Avg} \\
\midrule
ACIDS & $80.9$ & $48.2$  & $50.5$ & $\mathbf{56.7}$ & $59.1$ \\
\midrule
$\mathbf{PCD}$ & $\mathbf{92.6} $ & $\mathbf{49.7} $ & $\mathbf{56.7} $ & $53.4 $  & $\mathbf{63.4} $ \\
\bottomrule
\end{tabular}}
\label{tab:acids_comparison}
\end{table}

\begin{table}[htp!]
\vspace{-4mm}
\centering
 \caption{Clustering accuracy $(\%)$ on different initialization strategies for ResNet-50-based methods (supervised pre-training).} \resizebox{0.8\textwidth}{!}{
 \begin{tabular}{c|cccc|c|}
\toprule  Settings & \multicolumn{5}{c|}{PACS}  \\
\cmidrule(r){2-6} 
\multicolumn{1}{c}{} & \multicolumn{1}{|c}{$\mathcal{R} \rightarrow$ P} & \multicolumn{1}{c}{$\mathcal{R} \rightarrow$ A} & \multicolumn{1}{c}{$\mathcal{R} \rightarrow$ C} &
\multicolumn{1}{c}{$\mathcal{R} \rightarrow$ S} &
\multicolumn{1}{|c|}{Avg} \\
\midrule
Self-supervised pre-training  & $82.1$ & $53.4$  & $50.8$ & $43.6$ & $57.5$ \\
\midrule
Supervised pre-training & $\mathbf{93.3} $ & $\mathbf{54.6} $ & $\mathbf{59.1} $ & $\mathbf{56.8}$  & $\mathbf{66.0} $ \\
\bottomrule
\end{tabular}}
\label{tab:resnet18_full}
\end{table}

\begin{table}[htp!]
\vspace{-4mm}
\centering
 \caption{Clustering accuracy $(\%)$ on different datasets for ResNet-50-based methods (self-supervised pre-training).} \resizebox{1.0\textwidth}{!}{\begin{tabular}{c|ccc|c|cccc|c|cccc|c}
\toprule  Settings & \multicolumn{4}{c|}{Office-31} & \multicolumn{5}{c|}{Office-Home} & \multicolumn{5}{c}{PACS}  \\
\cmidrule(r){2-5} \cmidrule(r){6-10} \cmidrule{11-15} 
\multicolumn{1}{c}{} & \multicolumn{1}{|c}{$\mathcal{R} \rightarrow$ A} & \multicolumn{1}{c}{$\mathcal{R} \rightarrow$ W} & \multicolumn{1}{c}{$\mathcal{R} \rightarrow$ D} & 
\multicolumn{1}{|c|}{Avg} & 
\multicolumn{1}{c}{$\mathcal{R} \rightarrow$ Ar} & \multicolumn{1}{c}{$\mathcal{R} \rightarrow$ Cl} & \multicolumn{1}{c}{$\mathcal{R} \rightarrow$ Pr} & \multicolumn{1}{c}{$\mathcal{R} \rightarrow$ Rw} & \multicolumn{1}{|c|}{Avg} &
\multicolumn{1}{c}{$\mathcal{R} \rightarrow$ P} & \multicolumn{1}{c}{$\mathcal{R} \rightarrow$ A}& \multicolumn{1}{c}{$\mathcal{R} \rightarrow$ C}& \multicolumn{1}{c}{$\mathcal{R} \rightarrow$ S} & \multicolumn{1}{|c}{Avg} \\

\midrule
PO & $13.5\pm{0.9} $ &  $16.7 \pm{0.3}$ & $19.3\pm{2.4}$ & $16.5$ & $10.5\pm{0.6}$ & $8.4\pm{0.2}$ & $10.5\pm{0.7}$ & $9.1\pm{0.9}$ & $9.6$ & $28.3\pm{7.4}$ & $22.9\pm{2.6}$ & $24.0\pm{1.2}$ & $29.1\pm{2.3}$ & $26.1$\\
SO & $19.0\pm{5.0}$ & $26.3\pm{2.0}$ & $27.5\pm{3.0}$ & $24.3$ & $18.3\pm{1.2}$ & $11.2\pm{0.3}$ & $16.4\pm{1.3}$ & $16.7\pm{1.7}$ & $15.7$ & $40.7\pm{12.2}$ & $25.0\pm{1.4}$ & $29.7\pm{5.7}$ & $35.0\pm{5.0}$ & $32.6$\\
TO & $31.6 \pm{1.8}$ & $34.3  \pm{4.3}$  & $33.7  \pm{2.8}$  & $33.2$ & $17.9  \pm{2.0}$ & $10.1  \pm{0.1}$ & $20.7  \pm{1.9}$ & $16.6 \pm{1.8}$ & $16.3$  & $80.4  \pm{6.8}$ & $51.9 \pm{2.4}$ & $44.8  \pm{1.3}$ & $32.8  \pm{1.3}$ & $52.5$  \\
AO & $33.3\pm{0.6}$ & $37.6\pm{5.3}$  & $41.9\pm{2.7}$  & $37.6$ & $21.7\pm{2.2}$ & $17.9\pm{0.8}$ & $20.8\pm{3.7}$ & $27.5\pm{3.4}$ & $22.0$ & $80.0\pm{3.8}$ & $38.9\pm{3.6}$ & $55.6\pm{3.4}$ & $43.5\pm{3.7}$ & $54.5$\\
\midrule
$\mathbf{PCD}$ & $\mathbf{37.8}\pm{1.5}$ & $\mathbf{48.2}\pm{5.4}$ & $\mathbf{51.0}\pm{4.8}$	& $\mathbf{45.6}$ & $\mathbf{23.8}\pm{0.9}$ & $\mathbf{18.4}\pm{0.4}$ & $\mathbf{30.6}\pm{1.5}$ & $\mathbf{27.6}\pm{1.2}$ & $\mathbf{25.1}$ & $\mathbf{82.1}\pm{4.0}$ & $\mathbf{53.4}\pm{4.4}$ & $\mathbf{50.8}\pm{4.5}$ & $\mathbf{43.6}\pm{4.7}$ & $\mathbf{57.5}$\\
\bottomrule
\end{tabular}}
\label{tab:resnet50_full}
\end{table}

\subsection{Model transfer setting}
\begin{table}[htp!]
\vspace{-4mm}
\centering
 \caption{Clustering accuracy $(\%)$ on Office-31 for different model transfer methods.} \resizebox{1.0\textwidth}{!}{\begin{tabular}{c|ccc|c|ccc|c|ccc|c}
\toprule  Settings & \multicolumn{4}{c|}{ViT-B/16 (ssl) $\rightarrow$ ResNet-50 (ssl)} & \multicolumn{4}{c|}{ViT-B/16 (ssl) $\rightarrow$ ResNet-50 (sup)} & \multicolumn{4}{c|}{ViT-B/16 (ssl) $\rightarrow$ ResNet-18 (sup)}   \\
\cmidrule(r){2-5} \cmidrule(r){6-9} \cmidrule(r){10-13}
\multicolumn{1}{c}{} & \multicolumn{1}{|c}{$\mathcal{R} \rightarrow$ A} & \multicolumn{1}{c}{$\mathcal{R} \rightarrow$ W} & \multicolumn{1}{c}{$\mathcal{R} \rightarrow$ D} & 
\multicolumn{1}{|c|}{Avg} & 
\multicolumn{1}{c}{$\mathcal{R} \rightarrow$ A} & \multicolumn{1}{c}{$\mathcal{R} \rightarrow$ W} & \multicolumn{1}{c}{$\mathcal{R} \rightarrow$ D} & 
\multicolumn{1}{|c}{Avg} &  
\multicolumn{1}{|c}{$\mathcal{R} \rightarrow$ A} & \multicolumn{1}{c}{$\mathcal{R} \rightarrow$ W} & \multicolumn{1}{c}{$\mathcal{R} \rightarrow$ D} & 
\multicolumn{1}{|c}{Avg} \\
\midrule
PO & $20.1\pm{0.2}/13.5\pm{0.9}$ &  $26.7\pm{0.8}/16.7\pm{0.3}$ & $27.2\pm{0.3}/19.3\pm{2.4}$ & $24.7/16.5$ & $20.1\pm{0.2}/15.7\pm{0.7}$ & $26.7\pm{0.8}/24.2\pm{3.5}$ & $27.2\pm{0.3}/18.8\pm{2.2}$ & $24.7/19.6$ & $20.1\pm{0.2}/14.1\pm{1.6}$ & $26.7\pm{0.8}/17.9\pm{2.0}$ & $27.2\pm{0.3}/18.3\pm{2.9} $ & $24.7/16.8$ \\
SO & $43.2\pm{5.0}$ & $46.4\pm{4.5}$ & $37.2\pm{9.0}$ & $42.3$ & $43.2\pm{5.0}$ & $46.4\pm{4.5}$ & $37.2\pm{9.0}$ & $42.3$ &  $43.2\pm{5.0}$ & $46.4\pm{4.5}$ & $37.2\pm{9.0}$ & $42.3\pm{}$\\
TO & $32.6\pm{1.8}$ & $34.3\pm{4.3}$ & $33.7\pm{2.8}$ & $33.5$ & $43.7\pm{1.6}$ & $55.8\pm{2.1}$ & $\mathbf{52.0}\pm{3.8}$ & $50.5$ &  $38.0\pm{3.2}$ & $45.3\pm{1.6}$ & $46.6\pm{1.5}$ & $43.3$ \\
AO & $50.6\pm{3.7}$ & $49.7\pm{4.0}$ & $36.4\pm{5.0}$ & $45.6$ & $52.5\pm{3.5}$ & $53.7\pm{1.9}$ & $44.2\pm{4.1}$ & $50.1$ & $53.0\pm{3.8}$ & $47.2\pm{3.3}$ & $43.9\pm{4.9}$ & $48.0$ \\
\midrule
$\mathbf{PCD}$ & $\mathbf{51.7} \pm{2.9}$ &  $\mathbf{51.7} \pm{2.0}$ & $\mathbf{41.8} \pm{3.2}$ & $\mathbf{48.4}$ & $\mathbf{54.4}\pm{2.4}$ & $\mathbf{60.8}\pm{2.1}$ & $49.2\pm{3.3}$ & $\mathbf{54.8}$ &  $\mathbf{54.6} \pm{2.5}$ & $\mathbf{53.6} \pm{5.4}$ & $\mathbf{46.7} \pm{4.8}$ & $\mathbf{51.6}$ \\
\bottomrule
\end{tabular}}
\label{tab:model_trans_full}
\end{table}

\subsection{Limited-data and cluster-imbalanced and setting}
% \begin{table}[htp!]
% \vspace{-4mm}
% \centering
%  \caption{Clustering accuracy $(\%)$ on sub-sampled version of Office-31 for Resnet-50-based methods (self-supervised pretraining).} \resizebox{0.65\textwidth}{!}{\begin{tabular}{c|cccccccc}
% \toprule  Settings & $\mathcal{R} \rightarrow$ sub-A & $\mathcal{R} \rightarrow$ sub-W & $\mathcal{R} \rightarrow$ sub-D  &   Avg \\
% \midrule
% PO & $14.6\pm{1.2}$ & $16.7\pm{1.4}$ & $21.5\pm{1.5}$ & $17.6$\\
% SO & $21.1\pm{3.0}$ & $32.5\pm{2.6}$ & $36.0\pm{3.2}$ &  $29.9$\\
% TO & $31.4\pm{3.0}$ & $41.9\pm{3.6}$ & $45.1\pm{3.1}$ & $39.5$\\
% AO & $34.7\pm{3.5}$ & $40.8\pm{3.9}$ & $43.9\pm{3.6}$ & $39.8$\\
% \midrule
% $\mathbf{PCD}$ & $\mathbf{37.8} \pm{3.6}$ &	$\mathbf{46.4} \pm{3.3}$	& $\mathbf{47.0} \pm{3.7}$	& 	$\mathbf{43.7}$\\
% \bottomrule
% \end{tabular}}
% \label{tab:msdaofficehome}
% \end{table}

\begin{table}[htp!]
\vspace{-4mm}
\centering
 \caption{Clustering accuracy $(\%)$ on sub-sampled versions of different datasets for ResNet-50-based methods (self-supervised pre-training).} \resizebox{1.0\textwidth}{!}{\begin{tabular}{c|ccc|c|cccc|c|cccc|c}
\toprule  Settings & \multicolumn{4}{c|}{Office-31} & \multicolumn{5}{c|}{Office-Home} & \multicolumn{5}{c}{PACS}  \\
\cmidrule(r){2-5} \cmidrule(r){6-10} \cmidrule{11-15} 
\multicolumn{1}{c}{} & \multicolumn{1}{|c}{$\mathcal{R} \rightarrow$ sub-A} & \multicolumn{1}{c}{$\mathcal{R} \rightarrow$ sub-W} & \multicolumn{1}{c}{$\mathcal{R} \rightarrow$ sub-D} & 
\multicolumn{1}{|c|}{Avg} & 
\multicolumn{1}{c}{$\mathcal{R} \rightarrow$ sub-Ar} & \multicolumn{1}{c}{$\mathcal{R} \rightarrow$ sub-Cl} & \multicolumn{1}{c}{$\mathcal{R} \rightarrow$ sub-Pr} & \multicolumn{1}{c}{$\mathcal{R} \rightarrow$ sub-Rw} & \multicolumn{1}{|c|}{Avg} &
\multicolumn{1}{c}{$\mathcal{R} \rightarrow$ sub-P} & \multicolumn{1}{c}{$\mathcal{R} \rightarrow$ sub-A}& \multicolumn{1}{c}{$\mathcal{R} \rightarrow$ sub-C}& \multicolumn{1}{c}{$\mathcal{R} \rightarrow$ sub-S} & \multicolumn{1}{|c}{Avg} \\

\midrule
PO & $14.6\pm{1.2}$ & $16.7\pm{1.4}$ & $21.5\pm{1.5}$ & $17.6$ & $12.5\pm{0.8}$ & $9.3\pm{0.8}$ & $14.4\pm{1.1}$ & $12.2\pm{1.3}$ & $12.1$ & $43.0\pm{8.9}$ & $32.3\pm{5.9}$ & $29.1\pm{1.1}$ & $39.8\pm{3.7}$ & 36.1\\
SO & $21.1\pm{3.0}$ & $32.5\pm{2.6}$ & $36.0\pm{3.2}$ &  $29.9$ & $26.2\pm{1.3}$ & $19.5\pm{0.3}$ & $23.9\pm{1.0}$ & $26.9\pm{1.1}$ & $24.1 $ & $54.8\pm{4.4}$ & $37.8\pm{4.2}$ & $41.0\pm{5.5}$ & $47.2\pm{6.8}$ & $45.2$
\\
TO & $31.4\pm{3.0}$ & $41.9\pm{3.6}$ & $45.1\pm{3.1}$ & $39.5$ & $21.6\pm{1.8}$ & $11.9\pm{1.0}$ & $28.7\pm{1.3}$ & $22.8\pm{5.3}$ & $21.2$ & $65.1\pm{2.8}$ & $46.4\pm{2.3}$ & $47.8\pm{5.4}$ & $40.9\pm{1.2}$ & $50.0$
 \\
AO & $34.7\pm{3.5}$ & $40.8\pm{3.9}$ & $43.9\pm{3.6}$ & $39.8$ & $28.1\pm{0.6}$ & $21.8\pm{0.6}$ & $30.3\pm{2.6}$ & $29.4\pm{2.1}$ & $27.4$ & $65.4\pm{5.4}$ & $43.2\pm{6.7}$ & $51.1\pm{1.8}$ & $\mathbf{43.4}\pm{2.3}$ & $50.7$
\\
\midrule
$\mathbf{PCD}$ & $\mathbf{37.8} \pm{3.6}$ &	$\mathbf{46.4} \pm{3.3}$	& $\mathbf{47.0} \pm{3.7}$	& $\mathbf{43.7}$ & $\mathbf{28.7} \pm{1.0}$ & $\mathbf{22.3} \pm{0.4}$ & $\mathbf{32.7} \pm{2.6}$ & $\mathbf{31.2} \pm{3.2}$ & $\mathbf{28.7}$ &$\mathbf{66.1} \pm{4.7}$ & $\mathbf{48.0} \pm{1.5}$ & $\mathbf{51.8} \pm{2.0}$ & $\mathbf{43.4} \pm{1.9}$ & $\mathbf{52.4} $
\\
\bottomrule
\end{tabular}}
\label{tab:subsample_full}
\end{table}

Due to space constraints, we provide additional results for the sub-sampled versions of all three datasets in the appendix. PCD again outperforms other alternative methods consistently. AO, on average, performs better than TO, meaning that knowledge from the source can benefit target training. Similarly, SO improves upon PO in all cases. PCD achieves higher clustering accuracy than AO ($1-4\%$), illustrating that target model refinement is crucial for PCD's success.

\subsection{Ablation study}

\begin{table}[!ht]
    \centering
    \caption{Full ablation study on Office-31 dataset.}
    \begin{tabular}{c|c|c|c|c|c|c}
    \toprule
        Settings & $\mathcal{R} \rightarrow$W & diff & $\mathcal{R} \rightarrow$ A & diff & $\mathcal{R} \rightarrow$ D & diff \\ \midrule
        Full & $60.0$ & $0.0$ & $46.8$ & $0.0$ & $57.8$ & $0.0$ \\ 
        w/o prototype clustering & $49.7$ & $-10.3$ & $43.2$ & $-3.6$ & $53.2$ & $-4.6$ \\ 
        w/o MI & $52.7$ & $-7.3$ & $39.1$ & $-7.7$ & $54.79$ & $-3.01$ \\ 
        w/o CutMix & $54.5$ & $-5.5$ & $46.1$ & $-0.7$ & $54.6$ & $-3.2$ \\ 
        w/o Temporal Ensemble & $58.3$ & $-1.7$ & $46.6$ & $-0.2$ & $57.3$ & $-0.5$ \\ 
        w/o model privacy & $63.1$ & $3.1$ & $49.2$ & $2.4$ & $61.4$ & $3.6$ \\ 
        pooled source & $57.8$ & $-2.2$ & $45.6$ & $-1.2$ & $57.0$ & $-0.8$ \\ \bottomrule
    \end{tabular}
\end{table}

\begin{table}[H]
 %\vspace{-3mm}
\centering
\caption{Clustering accuracy (\%) on the task $\mathcal{R}$ $\rightarrow$ W (Office-31) under different variants  (ResNet-18).}
 \resizebox{0.75\columnwidth}{!}{ \begin{tabular}{ccccccc}
\toprule
Full & w/o prototype clustering &
 w/o MI &  w/o CutMix &  w/o Temporal Ensemble & w/o model privacy & pooled source \\
\midrule
 $60.0 \pm{2.6} $ & $49.7 \pm{2.8}$& $52.7\pm{3.0} $ & $54.5 \pm{5.2}$ & $58.3\pm{2.4}$ &$63.1 \pm{2.1}$ & $57.8 \pm{2.5}$ \\
\bottomrule
\end{tabular}}
\label{tab:ablation_full}
\end{table}

\section{Sensitivity plot}
\label{appendix:sensitivity}
\begin{figure}[htp!]
    \centering
        \includegraphics[width=1\textwidth]{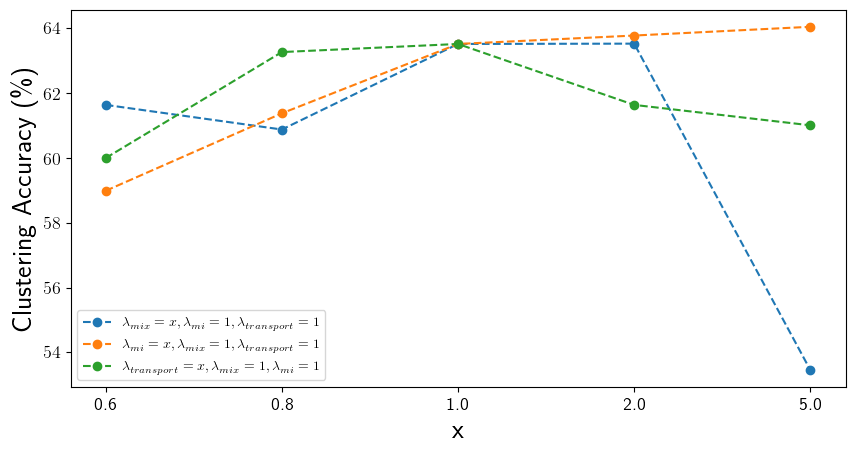}%
        \caption{Sensitivity plot for the coefficient of the losses. We fix the coefficient of the two losses to $1.0$ while varying the third loss from $0.6$ to $5.0$ and plot the clustering accuracy on the target data.
       }
        
        \label{fig:sensitivity_plot}%
\end{figure}

In Figure \ref{fig:sensitivity_plot}, we plot the sensitivity of the target clustering accuracy when we vary the coefficient in front of the loss. We can see that our method is not sensitive to different values of the coefficients except for when the $\lambda_{mix}$ coefficient is set to $5$. This result is expected since the $\lambda_{mix}$ is used as a regularization term and should not be set too high. We also observe that the performance can get even better via oracle validation by setting the $\lambda_{mi}$ to $2$ or $5$. However, we set the coefficient to $1$ for all three losses for all experiments.

\section{Running time and parameter size}
\label{appendix:param_size}

\begin{table}[htp!]
 %\vspace{-3mm}
\centering
\caption{Number of parameters and average running time per step for different clustering approaches for ResNet-18-based models.}
 \resizebox{0.8\columnwidth}{!}{ \begin{tabular}{ccc}
\toprule
Methods & Parameter size (millions) & Running time (s/step) \\
\midrule
ACIDS & 11.94 M & head1 - 0.52 s/step / head2 - 0.44 s/step\\
PCD & 11.32 M & 0.16 s/step \\
\bottomrule
\end{tabular}}
\label{tab:param_size}
\end{table}

\begin{table}[htp!]
 %\vspace{-3mm}
\centering
\caption{Average running time per step for different clustering approaches for ResNet-18-based models.}
 \resizebox{0.6\columnwidth}{!}{ \begin{tabular}{cc}
\toprule
ACIDS & PCD \\
\midrule
 head1 - 0.52 s/step / head2 - 0.44 s/step  & 0.16 s/step  \\
\bottomrule
\end{tabular}}
\label{tab:running_time}
\end{table}

\section{Connection with DeepCluster and SeLa}
\label{appendix:connection_deepcluster}
\citet{caron2018deep} propose DeepCluster to perform clustering and representation learning simultaneously. This method alternates between K-means for clustering and cross-entropy minimization for representation learning. While compatible with deep learning frameworks, the approach does have an obvious degenerate solution where all the samples get assigned to one cluster, yielding a constant representation. To overcome this, \citet{asano2019self} invent SeLa, which is similar to DeepCluster in the cross-entropy minimization step but differs from it in the pseudo-label assignment step. The authors explain that solving the K-means problem with equal partitioning constraints can avoid the degenerate solution. \citet{asano2019self} further recognize this as an instance of an optimal transport problem. Our clustering method is similar to SeLa in that we also solve the optimal transport problem during the pseudo-label assignment step. Unlike SeLa, we do not use the simplistic assumption that each cluster contains an equal number of data points. Instead, we dynamically update the cluster proportions using the predicted cluster probabilities. We also offer the interpretation of our method from the distribution alignment perspective. Moreover, our method is designed specifically for multi-domain data, and we also explore the use of our framework under the domain shift scenario.

\section{Full implementation details}
\label{appendix:implementation}
We follow the standard protocols for source-free unsupervised domain adaptation \citep{liang2020we}. Specifically, we use mini-batch SGD with a momentum of $0.9$ and weight decay of $0.001$. Both source and target encoders are initialized with ImageNet pre-trained networks \citep{russakovsky2015imagenet}, but the prototypes are initialized with a random linear layer. The initial learning rates are set to $0.001$ for the pre-trained encoders and $0.01$ for the randomly initialized layer. The learning rates, $\eta$, follows the following schedule: $\eta=\eta_0(1+10p)^{-0.75}$ where $\eta_0$ is the initial learning rate. We use the batch size of $64$ in both source and target learning. The initial value $\beta_0$ to learn domain-specific proportions is set to $0.9999$ for source clustering and $0.99$ for target clustering in all settings. We set the entropic regularization parameter, $\epsilon$, to $0.01$. The concentration parameter, $\alpha$, in the CutMix loss is set to $0.3$. The temporal ensemble coefficient, $\tau$, is equal to $0.6$. The source model and hyper-parameters are selected using the validation set of the source domain. The target model is trained using all the target data. We run our method with three different random seeds to calculate the standard deviation. We implement our method in PyTorch.

\section{Pseudo-code}
\label{appendix:algorithm}
\begin{algorithm}[htb!]
 \caption{Pseudo code for our framework.}
 \label{alg:pseudo_code}
\begin{algorithmic}
 \STATE {\bfseries 1. Source model training}
 \STATE {\bfseries Input:} source data - $\mathcal{X}^s = \{\mathcal{X}^s_d\}_{d=1}^D$, source model - $G^s=C_{\muv^s}(F_{\thetav^s}(.))$ (a randomly initialized $C_{\muv^s}$ and a pre-trained $F_{\thetav^s}$)
  \STATE {\bfseries Output:} updated $\thetav^s$, $\muv^s$

 \FOR{$t=1$ {\bfseries to} $T$}
 \STATE $\bullet$ Sample a mini-batch of source data
 \STATE $\bullet$ Update the proportions $B$ with Eq. \eqref{eq:proportions_update}
 \STATE $\bullet$ Solve the optimal transport problem in Eq. \eqref{eq:transport_problem} to obtain the transport map for each domain
 \STATE $\bullet$ Update the encoder and prototypes using Eq. \eqref{eq:source_joint_loss} with the transport map from the previous step
 \ENDFOR

  \STATE {\bfseries 2. Target model clustering}
 \STATE {\bfseries Input:} target data - $\mathcal{X}^t = \{\xv_j^{t} \}_{j=1}^{n_t}$, cluster labels from the source model - $G^s(x^t)$, target model - $G^t=C_{\muv^t}(F_{\thetav^t}(.))$ (a randomly initialized $C_{\muv^t}$ and a pre-trained $F_{\thetav^t}$)
 \STATE {\bfseries Output:} updated $\thetav^t$, $\muv^t$

  \FOR{$t=1$ {\bfseries to} $T$}
 \STATE $\bullet$ Sample a mini-batch of target data
 \STATE $\bullet$ Refine the hard-label with label smoothing and temporal ensemble
 \STATE $\bullet$ Update the the target model with the loss in Eq. \eqref{eq:target_init_loss}
 \ENDFOR

   \STATE {\bfseries 3. Target model refinement}
 \STATE {\bfseries Input:} $\mathcal{X}^t = \{\xv_j^{t} \}_{j=1}^{n_t}$, target model - $G^t$ ($C_{\muv^t}$ and $F_{\thetav^t}$ from step 2's output)
 \STATE {\bfseries Output:} updated $\thetav^t$, $\muv^t$
 
   \FOR{$t=1$ {\bfseries to} $T$}
 \STATE $\bullet$ Sample a mini-batch of target data
 \STATE $\bullet$ Update the proportions $B$ with Eq. \eqref{eq:proportions_update} 
 \STATE $\bullet$ Solve the optimal transport problem in Eq. \eqref{eq:transport_problem} to obtain the transport map for the target domain
 \STATE $\bullet$ Update the encoder and prototype using Eq. \eqref{eq:target_refinement} with the transport map from the previous step
 \ENDFOR

\end{algorithmic}
\end{algorithm}

\end{document}